\documentclass{article}

\PassOptionsToPackage{table}{xcolor}
\usepackage[table]{xcolor}

\usepackage[final]{corl_2026}

\usepackage[T1]{fontenc}
\usepackage{times}
\usepackage{amsmath}
\usepackage{amssymb}
\usepackage{mathtools}
\usepackage{bm}
\usepackage{dsfont}
\usepackage{esvect}
\usepackage{stmaryrd}
\usepackage{systeme}
\usepackage{orcidlink}

\usepackage{graphicx}
\usepackage{subcaption}
\usepackage[font={small}]{caption}
\usepackage[export]{adjustbox}
\usepackage{array}
\usepackage{booktabs}
\usepackage{multirow}
\usepackage{makecell}
\usepackage{float}

\usepackage{paralist}

\usepackage{enumitem}
\usepackage[normalem]{ulem}
\usepackage{soul}
\usepackage{xspace}
\usepackage{url}
\usepackage[nolist]{acronym}

\usepackage[most]{tcolorbox}
\usepackage{listings}
\usepackage{wrapfig}

\usepackage{tikz}
\usetikzlibrary{tikzmark}
\usetikzlibrary{calc}

\usepackage{outline}
\usepackage{siunitx}
\usepackage{stackengine}
\usepackage{scalerel}
\usepackage{todonotes}

\usepackage{cleveref}

\def\ours{SPAN-Nav\xspace}

\definecolor{rowcolor}{rgb}{0.898,0.949,0.969}

\def\ours{SPAN-Nav\xspace}

\definecolor{revone}{RGB}{214, 96, 77}   
\definecolor{revtwo}{RGB}{64, 139, 87}   
\definecolor{revthree}{RGB}{72, 118, 180} 
\definecolor{accolor}{RGB}{196, 138, 47} 

\definecolor{allrevcolor}{RGB}{128, 90, 160} 

\usepackage[linesnumbered,ruled,vlined]{algorithm2e}

\SetNlSty{textbf}{\color{gray}}{}   
\SetAlgoNlRelativeSize{-1}
\SetNlSkip{0.6em}

\usepackage{xcolor}

\definecolor{spanGreen}{HTML}{5F9A3D}
\definecolor{spanTeal}{HTML}{2B8F96}
\definecolor{spanBlue}{HTML}{4D98D5}
\definecolor{spanDarkTeal}{HTML}{2F6F7E}

\AtBeginDocument{
  \setlength{\dblfloatsep}{2pt plus 1pt minus 1pt}
  \setlength{\dbltextfloatsep}{4pt plus 1pt minus 1pt}
}

\title{
\textcolor{spanGreen}{SP}\textcolor{spanTeal}{A}\textcolor{spanBlue}{N}-Nav: 
Generalized 
\textcolor{spanGreen}{Sp}atial 
\textcolor{spanTeal}{A}ware\textcolor{spanBlue}{n}ess 
for Embodied Navigation
}

\begin{document}

\author{%
    \begin{minipage}{\dimexpr\linewidth-2\tabcolsep\relax}
    \centering
    \vspace{8pt}
	Jiahang Liu\orcidlink{0009-0002-3334-6692}$^{1,5,*}$ ~
	Tianyu Xu\orcidlink{0009-0003-5653-9902}$^{2,5,*}$ ~
    Jiawei Chen\orcidlink{0009-0001-2821-3484}$^{2,5,*}$ ~
	Lu Yue\orcidlink{0009-0006-8946-2704}$^{3,5,*}$ \\
    \vspace{2pt}
    Jiazhao Zhang\orcidlink{0000-0001-9459-293X}$^{1,5,*}$~
    Zhiyong Wang\orcidlink{0009-0008-4763-5288}$^{5,*}$ ~
    Minghan Li\orcidlink{0009-0009-5153-2406}$^{5}$ ~
    Qisheng Zhao\orcidlink{0009-0006-8540-5024}$^{3,5}$ \\
    \vspace{2pt}
    Anqi Li\orcidlink{0009-0001-0368-612X}$^{4,5}$ ~
    Qi Su\orcidlink{0009-0003-3945-0922}$^{1,5}$ ~
    Zhizheng Zhang\orcidlink{0000-0002-5360-7565}$^{5,\dagger}$ ~
    He Wang\orcidlink{0000-0002-3365-4620}$^{1,5,\dagger}$
    \vspace{3pt}
    \end{minipage}
	\\ 
    \begin{minipage}{\dimexpr\linewidth-2\tabcolsep\relax}
    \centering
    \normalsize
    \begin{tabular}{@{}c@{}}
    $^1$School of Computer Science\\
    $^2$School of Electronics Engineering and Computer Science\\
    $^3$School of Advanced Manufacturing and Robotics\quad $^4$Yuanpei College\\
    $^{1\text{--}4}$Peking University, Beijing, China\quad $^5$Galbot, Beijing, China\\[3pt]
    $^{*}$Equal contribution\quad $^{\dagger}$Corresponding authors
    \end{tabular}
    \end{minipage}
    \\
    \begin{minipage}{\dimexpr\linewidth-2\tabcolsep\relax}
    \centering
    \begin{tabular}{@{}c@{}}
    Project Page: \url{https://pku-epic.github.io/SPAN-Nav-Web/}
    \end{tabular}
    \end{minipage}
}

\maketitle

\begin{abstract}
Recent embodied navigation methods leveraging Vision-Language Models (VLMs) have shown strong generalization. 
However, reliable path planning in complex environments remains challenging due to insufficient spatial awareness.
The key challenge lies in learning spatial awareness that generalizes across diverse scenes and supports cross-task navigation reasoning.
Motivated by this challenge, we propose \ours, which advances generalized 3D spatial awareness for embodied navigation by estimating egocentric 3D occupancy from RGB video streams and using it as an intermediate spatial representation for navigation reasoning.
\ours represents 3D occupancy in a continuous latent space and projects it into a compact spatial token for VLM integration. This design provides navigation-relevant geometric cues while avoiding the inference overhead of dense volumetric representations.
To align occupancy perception with navigation reasoning, \ours adopts Two-Stage Spatial-to-Action Alignment (S2A-Align), a training strategy that first grounds navigation decisions on ground-truth occupancy and then adapts action reasoning to self-predicted spatial priors to reduce the training-inference gap.
Based on a large-scale dataset with 4.2M occupancy annotations across diverse scenes and navigation tasks, \ours achieves state-of-the-art performance on Vision-and-Language Navigation (VLN), urban navigation and point-goal navigation benchmarks, improving Success Rate (SR) by 4.6\%
/ 5.3\% on R2R-CE~/~RxR-CE, reducing MetaUrban collision cost by over 4$\times$, and reaching 90.9\% / 91.0\% zero-shot SR on InternVLA-N1 System-1 point-goal navigation benchmark.
Real-world experiments further demonstrate robust obstacle avoidance and reliable execution in cluttered environments.

\end{abstract}


\section{Introduction}

Embodied navigation is a fundamental capability for intelligent agents, requiring robust physical environment perception and safe, goal-directed action prediction~\cite{zhang2025embodied,zhang2024uni,zhang2024navid}.
Recent advanced embodied navigation methods leverage Vision-Language Models (VLMs), which provide strong cross-modal representations and enable promising generalization through large-scale pretraining and end-to-end action learning~\cite{zhang2025embodied,zhang2024uni,zhang2024navid,wang2025trackvla,xu2025mm,wang2025internvla}.
However, existing methods mainly focus on semantic vision-language grounding and lack critical spatial awareness for action planning, limiting their collision avoidance and feasible route planning in cluttered and structurally complex environments.


Among existing spatial representations, 3D occupancy provides an intermediate representation, offering structured spatial priors while reducing the influence of appearance-level variations.
Previous works~\cite{huang2023tri,zhang2025occupancy,liu2024let,liu2025occvla,xu2025occ} have demonstrated the effectiveness of occupancy representations in 3D spatial perception and planning.
However, most existing occupancy-based methods are developed for task-specific settings or relatively structured domains such as autonomous driving~\cite{zheng2024occworld,wang2024occsora,wei2024occllama}, making them difficult to directly transfer to embodied navigation.
This transfer difficulty mainly stems from the high variability of navigation environments, where obstacle density, spatial clutter, and layout structure differ substantially, challenging the learning of a unified 3D occupancy representation that generalizes across environments and supports versatile navigation tasks.

To address these challenges, we introduce \ours, an embodied navigation model with general spatial awareness, which estimates egocentric 3D occupancy from RGB video streams.
Nevertheless, directly reasoning over dense occupancy maps is computationally expensive for large VLM backbones.
To enable efficient integration with VLMs, \ours encodes the occupancy in a continuous latent space and projects it into a compact spatial token, learning a lightweight representation of surrounding 3D free-occupied structures that transfer across diverse scenes.
The spatial token provides navigation-related geometric cues, including traversability and obstacle layouts, guiding action reasoning toward feasible trajectories with reduced collision risks.




We further introduce Two-Stage Spatial-to-Action Alignment (S2A-Align) training strategy to effectively connect spatial perception with navigation reasoning.
In the first stage, the policy learns to ground navigation decisions on ground-truth occupancy priors, while in the second stage, it is adapted to plan from its own predicted spatial tokens, reducing the training-inference discrepancy.
To support generalizable spatial learning, we build a large-scale dataset of 4.2 million occupancy annotations spanning indoor and outdoor environments~\cite{ramakrishnan2021habitat,wu2024metaurban,miao2025towards} and diverse navigation tasks, including Vision-and-Language Navigation (VLN)~\cite{zhang2025embodied,zhang2024uni,zhang2024navid}, urban navigation~\cite{li2025urbanvla}, and collision-free point-goal navigation~\cite{xu2025mm}.



Extensive experiments demonstrate that SPAN-Nav establishes state-of-the-art performance across diverse benchmarks, ranging from high-level instruction following to low-level obstacle avoidance in both indoor and outdoor settings. 
Specifically, on VLN-RxR-CE \cite{ku2020room}, SPAN-Nav improves the Success Rate (SR) by \textbf{5.3\%}; on MetaUrban \cite{wu2024metaurban}, it achieves a \textbf{4$\times$} reduction in cumulative cost; and in home environments of InternVLA-N1 System-1 point-goal navigation benchmark~\cite{wang2025internvla}, it yields a \textbf{30.9\%} higher SR.
Notably, even when trained without VLN-specific occupancy annotations, \ours still improves SR on RxR-CE task by \textbf{2.1\%}, demonstrating its ability to transfer spatial awareness learned from other navigation tasks.
Real-world deployment further validates this robustness, demonstrating high task completion rates and effective obstacle avoidance in challenging real-world scenes with glass surfaces, narrow passages, dense obstacles, and other spatial constraints.

\begin{figure}[t]
\vspace{-1em}
  \centering
  \includegraphics[width=1.0\textwidth]{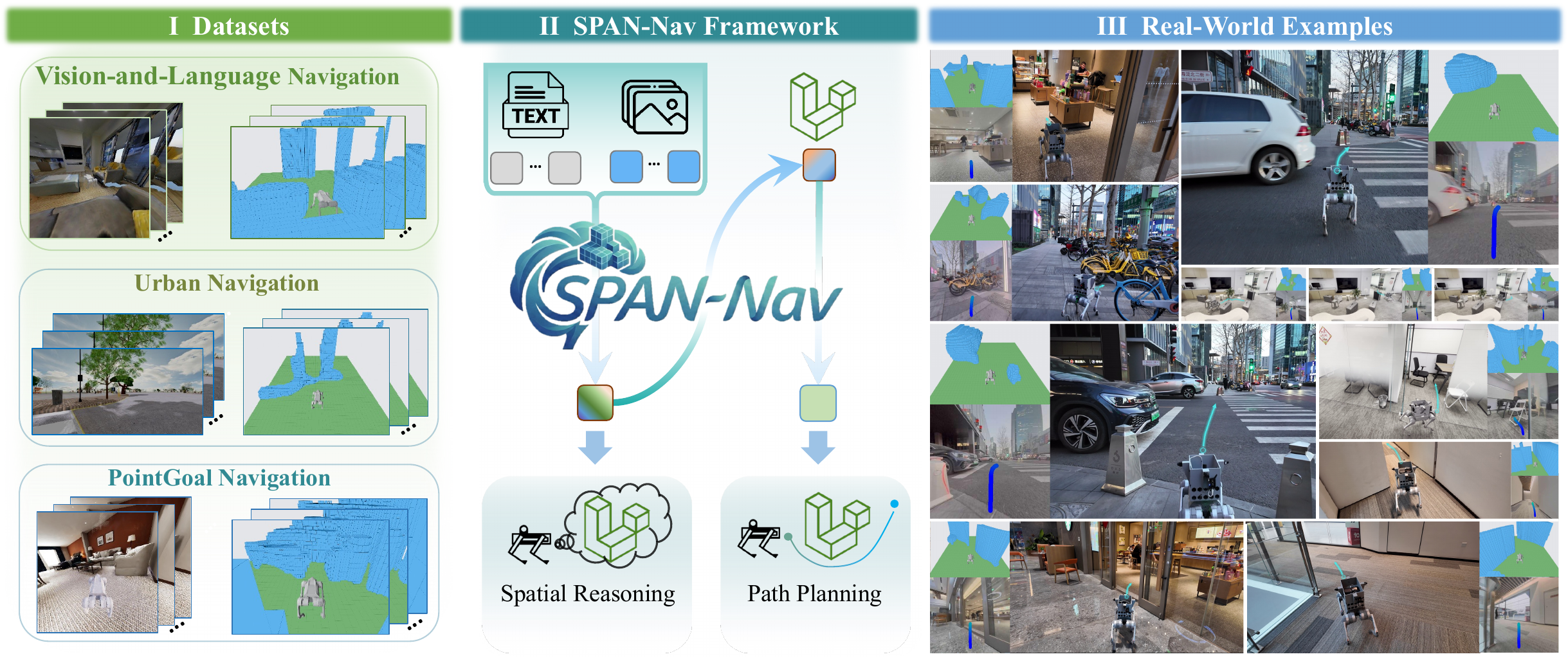}
  \caption{
    SPAN-Nav empowers end-to-end embodied navigation with generalized 3D spatial perception. By leveraging occupancy-guided spatial priors, SPAN-Nav enables reliable and safe navigation across versatile tasks and diverse, complex environments.
    }

  \label{fig:teaser}
  
\end{figure}

\section{Related Works}

\noindent\textbf{VLMs for Embodied Navigation.}
Embodied navigation has recently benefited from large-scale vision-language models and end-to-end visual-action learning. Early methods \citep{zhou2023navgpt, shah2022lmnav, qiao2023march,pan2023langnav,long2023discuss,lin2025navcot,chen2024mapgpt,qiao2025open} use off-the-shelf LLMs for instruction parsing, structured reasoning, or modular planning, but often convert visual observations into sparse language descriptions, limiting fine-grained spatial understanding. More recent methods \citep{zhang2024uni,zhang2024navid,wang2025trackvla,cheng2024navila, wei2025streamvln,zhou2025navgpt2, zheng2024towards, zhang-etal-2025-mapnav} fine-tune image- or video-based VLMs directly on navigation trajectories, enabling tighter visual-action coupling. Cross-task navigation models\citep{zhang2024uni,zhang2025embodied} further show that jointly learning from diverse navigation tasks improves generalization across scenarios. However, most VLM-based navigation policies still rely primarily on RGB observations and implicit visual-action correlations, making reliable planning difficult in cluttered scenes where traversability, obstacle layout, and local geometry are crucial.

\noindent\textbf{Spatial Awareness for Navigation Reasoning.}
Explicit spatial representations, such as depth, LiDAR, and 3D occupancy, have long supported navigation and planning~\citep{oleynikova2017voxblox,wijmans2019dd,chaplot2020learning,chaplot2020object}. 
BEVBert~\citep{an2023bevbert} and VLMaps~\citep{huang2023visual} explicitly ground navigation in map-based geometry, using depth to construct spatial maps. Thus, explicit spatial awareness itself is not new. Our distinction is an RGB-only, online, decodable egocentric occupancy latent compressed into one continuous spatial token and aligned with action reasoning through S2A-Align.
Among them, 3D occupancy explicitly models local free and occupied space, making it well suited for obstacle avoidance and feasible route estimation.
Prior works have studied occupancy prediction from RGB or depth inputs~\citep{Cao2022Monoscene,Wei2023Surroundocc,Tian2023Occ3D,Wang2023OpenOccupancy,Li2023FBOcc,Zhang2023OccFormer}, and recent occupancy world models or occupancy-augmented large models further use it for future prediction and action generation~\citep{zhang2025occupancy,liu2025occvla,xu2025occ,zheng2024occworld,wang2024occsora,wei2024occllama}.
However, these methods mostly target structured domains or task-specific settings such as autonomous driving, and their dense or discretized occupancy representations are costly to integrate into VLM-based navigation policies.
In contrast, \ours encodes the egocentric 3D occupancy into a continuous latent space and projects it into a compact spatial token for navigation reasoning across diverse tasks and scenes.

\begin{figure*}[!t]
\vspace{-1em}
  \centering
  \includegraphics[width=1.\textwidth]{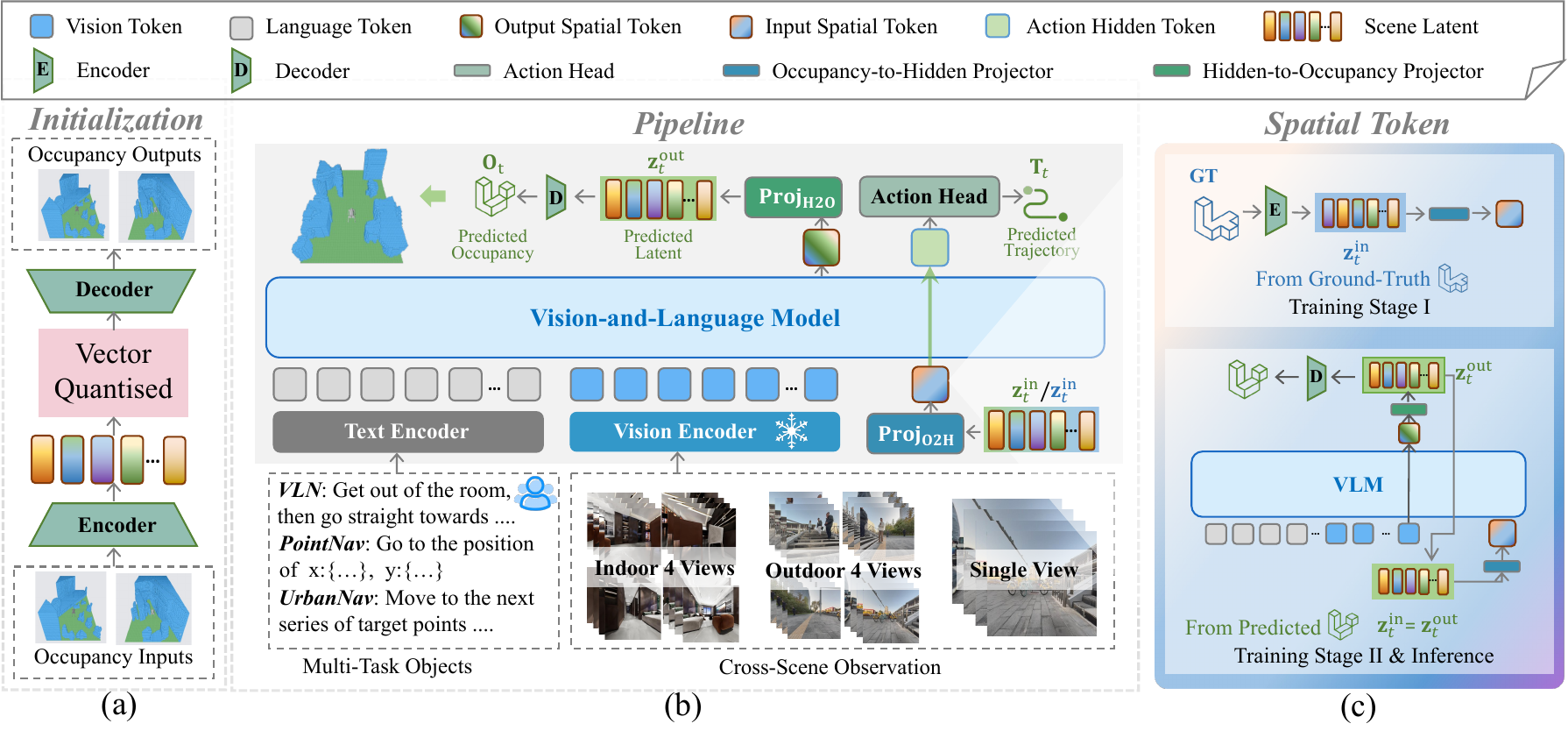}
  \caption{
  \textbf{Overview of SPAN-Nav.}
    (a) We pre-train a VQ-VAE-based encoder--decoder on cross-scene occupancy datasets for spatial representation initialization.
    (b) SPAN-Nav injects a spatial token into a VLM-based navigation pipeline for spatially aware decision-making.
    (c) Two-Stage Spatial-to-Action Alignment (S2A-Align): the spatial token is derived from ground-truth occupancy during Training Stage I, and from self-prediction during Training Stage II and inference.
  }
  \label{fig:main}
  
\end{figure*}

\section{Overview}


We formulate embodied navigation as a sequential decision-making problem, where an agent predicts a local trajectory $\mathbf{T}_t=\{\mathbf{a}_{m}\}_{m=1}^M$ from a language instruction $\mathbf{L}$ and RGB observation history $\mathbf{V}_t=\{\mathbf{v}_k\}_{k=1}^t$, with each pose $\mathbf{a}_{m}=(x,y,\theta)\in\mathbb{R}^3$ denoting an egocentric displacement and heading change over the planning horizon $M$. Built upon Qwen3-VL~\citep{Qwen3-VL}, \ours encodes instructions into language features $\mathbf{E}^{\text{L}}$ and processes multi-view RGB observations with the vision encoder, cross-modal projector, and spatiotemporal compression~\citep{zhang2025embodied} to obtain visual input $\mathbf{E}^{\text{V}}_t$. These features are fed into the VLM backbone to learn a compact spatial token under cross-scene occupancy supervision, which acts as a spatial prior for action reasoning and supports spatially aware trajectory prediction; the coordinated training strategy is detailed in \Cref{sec:train}.

\section{Model of \ours}

\subsection{Observation Encoding}\label{sec:obs}
Here, we detail the pipeline for processing RGB observations. The vision encoder from Qwen3-VL~\citep{Qwen3-VL} is employed to extract visual features from the observation sequence $\mathbf{V}_t$. Since continuous navigation generates an excessive volume of frames and features, we implement a spatiotemporal compression strategy from NavFoM~\citep{zhang2025embodied} to manage this redundancy. Specifically, historical frames are probabilistically sampled to a fixed number and spatially aggregated via average pooling. The resulting compressed history features are concatenated with intact current-frame features and projected into the latent space as the final visual input $\mathbf{E}^{\text{V}}_t$.

\subsection{Cross-Scene Spatial Token Learning}\label{sec:spa-tok}

To enable robust spatial awareness across diverse environments, we learn a compact spatial representation from occupancy prediction and align it with the VLM hidden space for navigation reasoning.

\noindent\textbf{Encoder-Decoder Initialization.}
As shown in \Cref{fig:main}~(a), we initialize the occupancy encoder Enc$(\cdot)$ and decoder Dec$(\cdot)$ with a 51.33M-parameter VQ-VAE pretrained on 6.22M automatically collected occupancy-only samples, inspired by~\citep{zheng2024occworld,wang2024occsora,wei2024occllama,zhang2025occupancy,xu2025occ}. This pretraining corpus differs from the 4.2M occupancy-annotated navigation samples used for policy training. The encoder maps a binary grid $\mathbf{O}_t\in\{0,1\}^{120\times80\times20}$ to $\mathbf{z}_t=\mathrm{Enc}(\mathbf{O}_t)\in\mathbb{R}^{15\times10\times128}$; pretraining uses a 4,096-entry, 128-D codebook. The encoder and decoder remain trainable during both navigation stages. \Cref{sec:EDInit} gives the pretraining recipe.

\noindent\textbf{Compact Cross-Scene Spatial Token.}
Although VQ-VAE quantization is effective in structured scenes, we find that discrete codes perform poorly in embodied navigation due to the heterogeneity of indoor environments, causing information loss in diverse spatial layouts (\Cref{tab:ablation_1} line1). 
We therefore use continuous occupancy latent $\mathbf{z}_t$ to construct spatial token $\mathbf{h}_t$.
To align occupancy latents with VLM hidden states, we introduce two projectors. The occupancy-to-hidden projector $\text{Proj}_{\text{O2H}}(\cdot)$ compresses the input latent $\mathbf{z}^{\text{in}}_t$ into a single compact spatial token $\mathbf{h}^{\text{in}}_t$:
\begin{equation}
\mathbf{h}^{\text{in}}_t = \text{Proj}_{\text{O2H}}(\mathbf{z}^{\text{in}}_t).
\end{equation}
Conversely, the hidden-to-occupancy projector $\text{Proj}_{\text{H2O}}(\cdot)$ maps the VLM output spatial token $\mathbf{h}^{\text{out}}_t$ back to the occupancy latent space:
\begin{equation}\label{eq:spatial-out}
\mathbf{h}^{\text{out}}_t=\text{VLM}(\mathbf{E}^{\text{L}}, \mathbf{E}_t^{\text{V}}), \quad 
\mathbf{z}^{\text{out}}_t = \text{Proj}_{\text{H2O}}(\mathbf{h}^{\text{out}}_t).
\end{equation}
This design compresses high-dimensional occupancy into a lightweight spatial token while preserving continuous geometric variations across heterogeneous scenes. Depending on the training stage, $\mathbf{h}^{\text{in}}_t$ is derived from either ground-truth or self-predicted occupancy latents $\mathbf{z}^{\text{in}}_t$ (\Cref{sec:train}).

\noindent\textbf{Cross-Scene Occupancy Supervision.}
To endow the spatial token with explicit 3D awareness, we supervise both occupancy reconstruction and latent consistency. The predicted latent $\mathbf{z}^{\text{out}}_t$ is decoded into an occupancy grid:
$
    \hat{\mathbf{O}}_t = \text{Dec}(\mathbf{z}_t^{\text{out}}).
$
We optimize the reconstruction with cross-entropy and Lovász-softmax losses:
$\label{eq:loss-occ}
    \mathcal{L}_{\text{occ}} =
    \mathcal{L}_c(\mathbf{O}^{\text{GT}}_t, \hat{\mathbf{O}}_t) +
    \mathcal{L}_l(\mathbf{O}^{\text{GT}}_t, \hat{\mathbf{O}}_t),
$
where $\mathcal{L}_c$ and $\mathcal{L}_l$ denote the cross-entropy and Lovász-softmax losses, respectively. To preserve the occupancy latent manifold, we further impose:
$\label{eq:loss-latent}
    \mathcal{L}_{\text{latent}} =
    \| \mathrm{Enc}(\mathbf{O}^{\text{GT}}_t) - \mathbf{z}^{\text{out}}_t \|_2^2.
$
Together, these objectives encourage SPAN-Nav to infer accurate 3D geometry from 2D observations while maintaining high-fidelity spatial latents.

\subsection{Spatial-Prior-Guided Action Reasoning}\label{sec:cot}
To better integrate the extracted spatial information with action generation, we leverage the predicted occupancy latent embeddings as explicit spatial priors rather than treating occupancy prediction merely as an auxiliary task. 
Specifically, the occupancy latent $\mathbf{z}^{\text{out}}_t$ obtained from \Cref{eq:spatial-out} is projected into the VLM hidden space and injected into the action prediction process, enabling trajectory generation to be conditioned on both visual-language context and 3D environmental structure.
The generation process of the predicted trajectory $\mathbf{T}_t$ is formalized as follows:
\begin{equation}\label{eq:traj}
    \mathbf{E}^{\text{act}}_t = \text{VLM}(\mathbf{E}^{\text{L}}, \mathbf{E}^{\text{V}}_t, \text{Proj}_{\text{O2H}}(\mathbf{z}^{\text{out}}_t)), \quad 
    \mathbf{T}_t=\text{AH}(\mathbf{E}^{\text{act}}_t),
\end{equation}
where $\mathbf{E}^{\text{act}}_t$ represents the predicted action hidden state, and AH$(\cdot)$ denotes a multi-layer MLP for trajectory planning. 
We supervise trajectory prediction on large-scale cross-domain datasets by minimizing:
$
    \mathcal{L}_{\text{act}} = \| \mathbf{T}^{\text{GT}}_t - \mathbf{T}_t \|_2^2,
$\label{eq:loss-act}
where $\mathbf{T}^{\text{GT}}_t$ is the ground-truth trajectory.
By explicitly conditioning action generation on the learned spatial prior, the model is encouraged to focus on navigation-critical cues, such as traversable regions and nearby obstacles, while suppressing irrelevant volumetric details. We further visualize these navigation-critical cues in the ~\Cref{fig:sim-vis}.
In Stage II and inference, the two VLM forward in \Cref{eq:spatial-out,eq:traj} are consecutive autoregressive decoding steps sharing a KV cache after a single prefill of $(\mathbf{E}^{\mathrm{L}},\mathbf{E}^{\mathrm{V}}_t)$. The first predicts $\mathbf{h}^{\mathrm{out}}_t$, which is mapped through $\text{Proj}_{\mathrm{H2O}}$ and $\text{Proj}_{\mathrm{O2H}}$ to the input spatial token for the action step. The occupancy decoder supplies supervision and visualizations; it is not required for action inference. Stage I instead derives the input spatial token from ground-truth occupancy.


\begin{figure}[t]
  \centering

  \begin{minipage}{0.54\textwidth}
    \centering
    \includegraphics[width=\linewidth]{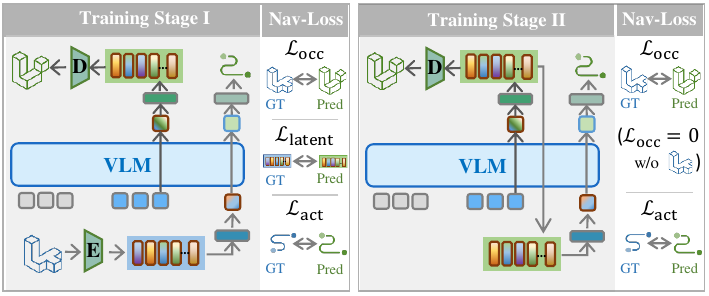}
    \caption{
      \textbf{Overview of the staged S2A-Align training process and the associated navigation task losses.}
    The first stage uses ground-truth occupancy, while the second stage transitions self-predicted spatial tokens.
    }
    \label{fig:stage}
  \end{minipage}
  \hfill
  \begin{minipage}{0.42\textwidth}
    \centering
    \includegraphics[width=\linewidth]{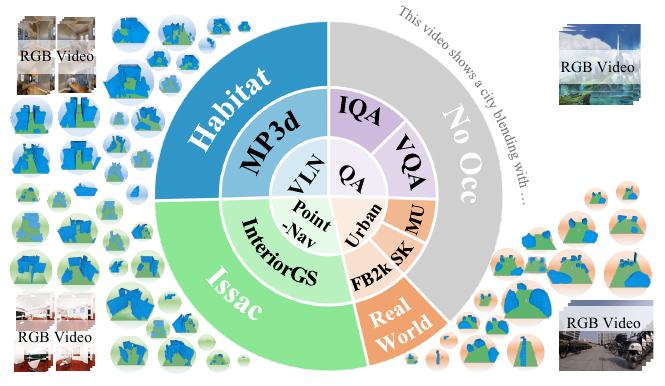}
    \caption{
      \textbf{Overview of the dataset composition.}
    MU, SK, and FB2K denote MetaUrban, SeKai, and FrodoBots-2K.
    }
    \label{fig:dataset}
  \end{minipage}

\end{figure}
\subsection{Two-Stage Spatial-to-Action Alignment}\label{sec:train}

For navigation training, we propose \textbf{Two-Stage Spatial-to-Action Alignment (S2A-Align)}, which progressively aligns spatial representations with action reasoning through two stages, as shown in \Cref{fig:stage}. 
The stage I establishes 3D spatial awareness and grounds action reasoning on clean occupancy representations, while the stage II further adapts the model to end-to-end navigation from visual observations.
During both stages, we co-train \ours with general QA data as auxiliary supervision. The QA loss is denoted as $\mathcal{L}_{\text{qa}}$, with details provided in Appendix \Cref{sec:QA}.

In the stage I, we train \ours on cross-scene and multi-task datasets with full occupancy annotations on noise-free spatial representations. The joint loss is:
\begin{equation}
    \mathcal{L}_{\text{s1}} = \mathcal{L}_{\text{occ}} + \mathcal{L}_{\text{latent}} + \mathcal{L}_{\text{act}} + \mathcal{L}_{\text{qa}},
\end{equation}
where $\mathcal{L}_{\text{occ}}$, $\mathcal{L}_{\text{latent}}$, and $\mathcal{L}_{\text{act}}$ denote the occupancy reconstruction, latent consistency, and action supervision losses in \Cref{eq:loss-occ,eq:loss-latent,eq:loss-act}. To reduce perception noise, action reasoning in \Cref{eq:traj} is conditioned on latent embeddings from ground-truth occupancy, i.e., $\mathbf{z}^{\text{in}}_t = \text{Enc}(\mathbf{O}_t^{\text{GT}})$.

Subsequently, in the stage II, we bridge the gap between training and inference by transitioning the trajectory planning process to rely on the model's self-predicted occupancy latent embeddings, i.e., $\mathbf{z}^{\text{in}}_t$ = $\mathbf{z}^{\text{out}}_t$ which from \Cref{eq:spatial-out}.
This stage is conducted on a mixed dataset comprising both occupancy-annotated and unannotated samples.
We formulate the optimization objective to balance two goals: maintaining high-fidelity spatial interpretation via the occupancy loss (\Cref{eq:loss-occ}), and adapting the policy to internal spatial reasoning via the action loss (\Cref{eq:loss-act}). The joint loss is :
\begin{equation}
    \mathcal{L}_{\text{s2}} = \mathcal{L}_{\text{occ}} + \mathcal{L}_{\text{act}} + \mathcal{L}_{\text{qa}},
\end{equation}
where $\mathcal{L}_{\text{occ}}$ is set to 0 for samples lacking ground-truth occupancy annotations.
This fine-tuning process empowers SPAN-Nav to prioritize navigation-critical spatial features over irrelevant volumetric noise while maintaining grounded spatial awareness, enabling robust end-to-end inference.

\section{Dataset of \ours} \label{section:data_set}

\subsection{Overview} \label{section:task_setting}


To train \ours, we construct a large-scale and diverse dataset spanning indoor and outdoor scenes, simulated and real-world environments, and multiple navigation tasks. 
It contains 7.08M samples, including 5.08M trajectories, 0.93M Image-QA samples, and 1.07M Video-QA samples. Among them, 4.2M samples are annotated with occupancy for spatial awareness learning.

The dataset spans three representative embodied navigation tasks. 
\noindent\textbf{Vision-Language Navigation (VLN)} requires agents to follow natural language instructions grounded in visual observations, emphasizing instruction following and visual reasoning. 
\noindent\textbf{Urban Navigation} requires agents to navigate complex urban environments toward instructed destinations, focusing on path planning and obstacle avoidance in realistic settings. 
\noindent\textbf{PointGoal Navigation} requires agents to reach coordinate-specified targets, emphasizing collision avoidance. 
Together, these tasks cover key navigation abilities, including instruction interpretation, path planning, and safe cross-scene navigation.

\subsection{Navigation Data Collection and Occupancy Annotation}


All navigation data follow a unified format, including egocentric videos, task instructions, ground-truth waypoints, and occupancy annotations. We use task-specific strategies to obtain accurate trajectories. \noindent\textbf{VLN (1.8M):} We collect R2R and RxR data in VLN-CE by recording RGB videos, instructions, and trajectories along ground-truth paths, and further augment training with DAgger data~\citep{ross2011reduction}. \noindent\textbf{Urban Navigation (1.28M):} We use MetaUrban for simulation data and Sekai~\citep{li2025sekai} and FrodoBots-2K~\citep{frodobots-2k-2025} for real-world data, where noisy trajectories are smoothed and lifted with HTL~\citep{li2025urbanvla} (Details in Appendix \Cref{sec:traj_for_urban}). \noindent\textbf{PointGoal Navigation (2M):} We sample start-goal pairs in 1,000 InteriorGS indoor scenes~\citep{miao2025physicallyexecutable3dgaussian} and generate smooth collision-free trajectories with an online \emph{A*+VO fusion} planner (Details in Appendix \Cref{sec:traj_for_point}).

To obtain large-scale occupancy annotations, we design separate pipelines for simulated and real-world data. \noindent\textbf{For annotation in simulation,} we convert 1,091 scenes from InteriorGS~\citep{InteriorGS2025} and MP3D~\citep{matterport3d} into an Isaac Sim-compatible format~\citep{NVIDIA_Isaac_Sim}, and use Isaac Sim to extract 3D occupancy at a voxel size of $0.05\,\mathrm{m} \times 0.05\,\mathrm{m} \times 0.1\,\mathrm{m}$. At each robot pose, we collect an agent-centered local occupancy map spanning 4\,m front, 2\,m rear, and 2\,m laterally.
\noindent\textbf{For annotation in real world,} due to the lack of complete scene geometry, we reconstruct FrodoBots-2K scenes~\citep{frodobots-2k-2025} using Depth Anything 3~\citep{depthanything3}, voxelize the resulting point clouds into occupied/free space, and following the same labeling strategy as in simulation for occupancy maps. See Appendix \Cref{sec:Occ_annotation} for details.

\section{Experiments}\label{sec:exp}

\begin{table}[t]
    \centering
    \caption{\textbf{Comparison on VLN-CE under RGB-only and multi-sensory (RGB-D + odometry) settings.}
$^{*}$ denotes methods using the waypoint predictor from~\citep{hong2022bridging}. \ours VLN-Occ-Free denotes that training without VLN occupancy annotations in both training stages.
Our RGB-only method achieves SOTA performance.
}
    \fontsize{6.2pt}{6.8pt}\selectfont
    \setlength{\tabcolsep}{4.2pt}
    \renewcommand{\arraystretch}{0.72}
    \setlength{\aboverulesep}{0.25ex}
    \setlength{\belowrulesep}{0.25ex}
    \resizebox{\linewidth}{!}{
    \begin{tabular}{l|ccc|cccc|cccc}
        \toprule
        \multirow{2}{*}{Method} 
        & \multicolumn{3}{c|}{Observation} 
        & \multicolumn{4}{c|}{R2R Val-Unseen} 
        & \multicolumn{4}{c}{RxR Val-Unseen} \\
        & RGB & Depth & Odo. 
        & NE $\downarrow$ & OS $\uparrow$ & SR $\uparrow$ & SPL $\uparrow$ 
        & NE $\downarrow$ & SR $\uparrow$ & SPL $\uparrow$ & nDTW $\uparrow$ \\
        \midrule

        CMA$^{*}$~\citep{hong2022bridging}  
        & \checkmark & \checkmark & \checkmark 
        & 6.20 & 52.0 & 41.0 & 36.0 
        & 8.76 & 26.5 & 22.1 & 47.0 \\

        VLN$\circlearrowright$BERT$^{*}$~\citep{hong2022bridging}  
        & \checkmark & \checkmark & \checkmark 
        & 5.74 & 53.0 & 44.0 & 39.0 
        & 8.98 & 27.0 & 22.6 & 46.7 \\

        AO-Planner~\citep{chen2024affordances}  
        & \checkmark & \checkmark &  
        & 5.55 & 59.0 & 47.0 & 33.0 
        & 7.06 & 43.3 & 30.5 & 50.1 \\

        Reborn$^{*}$~\citep{an20221st}  
        & \checkmark & \checkmark & \checkmark 
        & 5.40 & 57.0 & 50.0 & 46.0 
        & 5.98 & 48.6 & 42.0 & 63.3 \\

        ETPNav$^{*}$~\citep{an2024etpnav}  
        & \checkmark & \checkmark & \checkmark 
        & 4.71 & 65.0 & 57.0 & 49.0 
        & 5.64 & 54.7 & 44.8 & 61.9 \\

        HNR$^{*}$~\citep{wang2024lookahead}  
        & \checkmark & \checkmark & \checkmark 
        & 4.42 & 67.0 & 61.0 & 51.0 
        & 5.50 & 56.3 & 46.7 & 63.5 \\

        NaVid~\citep{zhang2024navid} 
        & \checkmark &  &  
        & 5.72 & 49.2 & 41.9 & 36.5 
        & 5.72 & 45.7 & 38.2 & - \\

        NaVILA~\citep{cheng2024navila} 
        & \checkmark &  &  
        & 5.22 & 62.5 & 54.0 & 49.0 
        & 6.77 & 49.3 & 44.0 & 58.8 \\

        Uni-NaVid~\citep{zhang2024uni} 
        & \checkmark &  &  
        & 5.58 & 53.3 & 47.0 & 42.7 
        & 6.24 & 48.7 & 40.9 & - \\

        JanusVLN~\citep{zeng2026janusvln} 
        & \checkmark &  &  
        & 4.78 & 65.2 & 60.5 & 56.8 
        & 6.06 & 56.2 & 47.5 & 62.1 \\

        DualVLN~\citep{wei2025dualvln} 
        & \checkmark &  &  
        & 4.05 & 70.7 & 64.3 & 58.5 
        & 4.58 & 61.4 & 51.8 & \textbf{70.0} \\

        Stream VLN~\citep{wei2025streamvln} 
        & \checkmark &  &  
        & 4.98 & 64.2 & 56.9 & 51.9 
        & 6.22 & 52.9 & 46.0 & 61.9 \\

        NavFom~\citep{zhang2025embodied} 
        & \checkmark &  &  
        & 4.61 & 72.1 & 61.7 & 55.3 
        & 4.74 & 64.4 & 56.2 & 65.8 \\

        \midrule

        \textbf{\ours VLN-Occ-Free} 
        & \checkmark &  &  
        & \textbf{4.02} & 73.1 & 63.3 & 57.3 
        & 4.50 & 66.5 & 57.0 & 67.3 \\

        \rowcolor{rowcolor}
        \textbf{\ours} 
        & \checkmark &  &  
        & 4.07 & \textbf{75.3} & \textbf{66.3} & \textbf{59.3} 
        & \textbf{4.20} & \textbf{69.7} & \textbf{60.1} & 67.9 \\

        \bottomrule
    \end{tabular}
    }
    \vspace{-5pt}
    \label{tab:r2r_rxr}
\end{table}


\subsection{Experiment Setup}

\noindent\textbf{Benchmarks.}\label{benchmark}
We evaluate \ours on three benchmarks: Vision-Language Navigation (VLN), Urban Navigation, and PointGoal Navigation. For VLN, we use the Val-UnSeen splits of R2R-CE~\citep{krantz2020beyond} and RxR-CE~\citep{ku2020room}. For Urban Navigation, we evaluate MetaUrban PointNav and SocialNav~\citep{wu2024metaurban} on 1,000 MetaUrban-test scenes and 100 unseen wheelchair-embodiment scenes. For PointGoal Navigation, we follow the InternVLA-N1(S1) setting~\citep{wang2025internvla,cai2025navdp}, using 40 InternScenes~\citep{zhong2025internscenes} scenes with the same 100 start-goal pairs and the Dingo omnidirectional robot.

\noindent\textbf{Metrics.}
We follow standard navigation metrics~\citep{anderson2018vision}, including Success Rate (SR), Oracle Success rate (OS), Success weighted by Path Length (SPL), cumulative Cost for collision avoidance (Cost), Social Navigation Score (SNS), normalized Dynamic Time Warping (nDTW), and Navigation Error (NE). Since success criteria vary across tasks, we adopt the default criteria of each benchmark.
For occupancy evaluation, decoded predictions are binarized at 0.5 and scored by mean occupied-voxel IoU.

\noindent\textbf{Training recipe.}
Stage I uses 2M occupancy-annotated trajectories and 0.5M QA samples; Stage II uses the full 7.08M navigation/QA dataset, with one epoch per stage. We use AdamW, a global batch size of 512, BF16, and cosine learning-rate scheduling with 3\% warmup. The vision tower is frozen; the vision merger uses learning rate $2\times10^{-4}$ and the VLM, action head, and occupancy modules use $2\times10^{-5}$. Two-stage training takes 12 hours on 32 H100 GPUs. VQ-VAE pretraining is additional (\Cref{sec:EDInit}).


\subsection{Performance Analysis}

\noindent\textbf{Benchmark Results.}
Across VLN, Urban Navigation, and PointGoal Navigation, \ours achieves state-of-the-art performance on all evaluated benchmarks. 
On VLN, as shown in \Cref{tab:r2r_rxr}, \ours outperforms the previous state-of-the-art NavFoM~\citep{zhang2025embodied}, improving SR by 4.6\% on R2R-CE and 5.3\% on RxR-CE, while using fewer training samples (7.08M vs. 12.7M). 
Notably, the variant trained without VLN occupancy annotations in both training stages called \textbf{\ours VLN-Occ-Free}, still surpasses prior methods despite a slight drop compared with the full model. 
This suggests that spatial awareness learned from UrbanNav and PointNav can generalize to VLN, highlighting the transferability of the learned occupancy-based spatial representation.
On MetaUrban, \ours consistently outperforms prior RL methods~\citep{schulman2017proximalpolicyoptimizationalgorithms,fujimoto2019benchmarkingbatchdeepreinforcement,sun2021safeexplorationsolvingearly,kostrikovoffline,fujimoto2021minimalistapproachofflinereinforcement}, imitation learning baselines~\citep{Bain1995AFF,ho2016generativeadversarialimitationlearning}, and VLA baselines~\citep{li2025urbanvla}. As reported in \Cref{tab:metaurban_benchmark}, \ours reduces cumulative Cost by over 4$\times$ compared with UrbanVLA while maintaining strong social compliance (SNS=0.96). 
For PointGoal Navigation, \ours also achieves substantial gains, improving SR by over 30.9\% in Home environments and 19.6\% in Commercial environments, as shown in \Cref{tab:sim_results}.

\begin{table*}[t]
    \centering
    \caption{\textbf{Comparison on the benchmark of PointNav and SocialNav tasks in the MetaUrban-12K dataset.} We compare our method with seven strong baselines with LiDAR observation. The \textbf{best} and the \underline{second best} results are denoted by \textbf{bold} and \underline{underline}, respectively.}
    \label{tab:metaurban_benchmark}
    \vspace{-5pt}
    \scriptsize
    \setlength{\tabcolsep}{3.2pt}
    \renewcommand{\arraystretch}{0.95}
    \resizebox{\textwidth}{!}{
    \begin{tabular}{l|c|ccc|ccc|ccc|ccc}
        \toprule
        \multirow{2}{*}{Method} 
        & \multirow{2}{*}{Observation} 
        & \multicolumn{3}{c|}{PointNav~(Test)} 
        & \multicolumn{3}{c|}{PointNav~(Unseen)} 
        & \multicolumn{3}{c|}{SocialNav~(Test)} 
        & \multicolumn{3}{c}{SocialNav~(Unseen)} \\
        & 
        & SR$\uparrow$ & SPL$\uparrow$ & Cost$\downarrow$ 
        & SR$\uparrow$ & SPL$\uparrow$ & Cost$\downarrow$ 
        & SR$\uparrow$ & SNS$\uparrow$ & Cost$\downarrow$ 
        & SR$\uparrow$ & SNS$\uparrow$ & Cost$\downarrow$ \\
        \midrule

        PPO~\citep{schulman2017proximalpolicyoptimizationalgorithms} 
        & LiDAR 
        & 66 & 64 & 0.51 
        & 49 & 45 & 0.78 
        & 34 & 64 & 0.66 
        & 24 & 57 & 0.51 \\

        PPO-Lag~\citep{fujimoto2019benchmarkingbatchdeepreinforcement} 
        & LiDAR 
        & 60 & 58 & \underline{0.41} 
        & 60 & 57 & \underline{0.53} 
        & 17 & 51 & 0.33 
        & 8 & 47 & \underline{0.50} \\

        PPO-ET~\citep{sun2021safeexplorationsolvingearly} 
        & LiDAR 
        & 57 & 53 & 0.47 
        & 53 & 49 & 0.65 
        & 5 & 52 & \underline{0.26} 
        & 2 & 50 & 0.62 \\

        IQL~\citep{kostrikovoffline} 
        & LiDAR 
        & 36 & 33 & 0.49 
        & 30 & 27 & 0.63 
        & 36 & 67 & 0.39 
        & 27 & 62 & 3.05 \\

        TD3+BC~\citep{fujimoto2021minimalistapproachofflinereinforcement} 
        & LiDAR 
        & 29 & 28 & 0.77 
        & 20 & 20 & 1.16 
        & 26 & 61 & 0.62 
        & 32 & 64 & 1.53 \\

        BC~\citep{Bain1995AFF} 
        & LiDAR 
        & 36 & 28 & 0.83 
        & 32 & 26 & 1.15 
        & 28 & 56 & 1.23 
        & 18 & 54 & 0.58 \\

        GAIL~\citep{ho2016generativeadversarialimitationlearning} 
        & LiDAR 
        & 47 & 36 & 1.05 
        & 40 & 32 & 1.46 
        & 34 & 63 & 0.71 
        & 28 & 61 & 0.67 \\

        UrbanVLA~\citep{li2025urbanvla} 
        & RGB 
        & \underline{94} & \underline{91} & 0.94 
        & \textbf{97} & \textbf{95} & 0.84 
        & \underline{91} & \underline{87} & 0.81 
        & \underline{88} & \underline{85} & 0.82 \\

        \rowcolor{rowcolor}
        \textbf{\ours} 
        & RGB 
        & \textbf{94} & \textbf{93} & \textbf{0.22} 
        & \underline{92} & \underline{91} & \textbf{0.20} 
        & \textbf{92} & \textbf{96} & \textbf{0.19} 
        & \textbf{93} & \textbf{96} & \textbf{0.28} \\

        \bottomrule
    \end{tabular}
    }
\end{table*}

\begin{figure*}[!t]
  \vspace*{-2.5mm}  
  \centering
  \includegraphics[width=\textwidth]{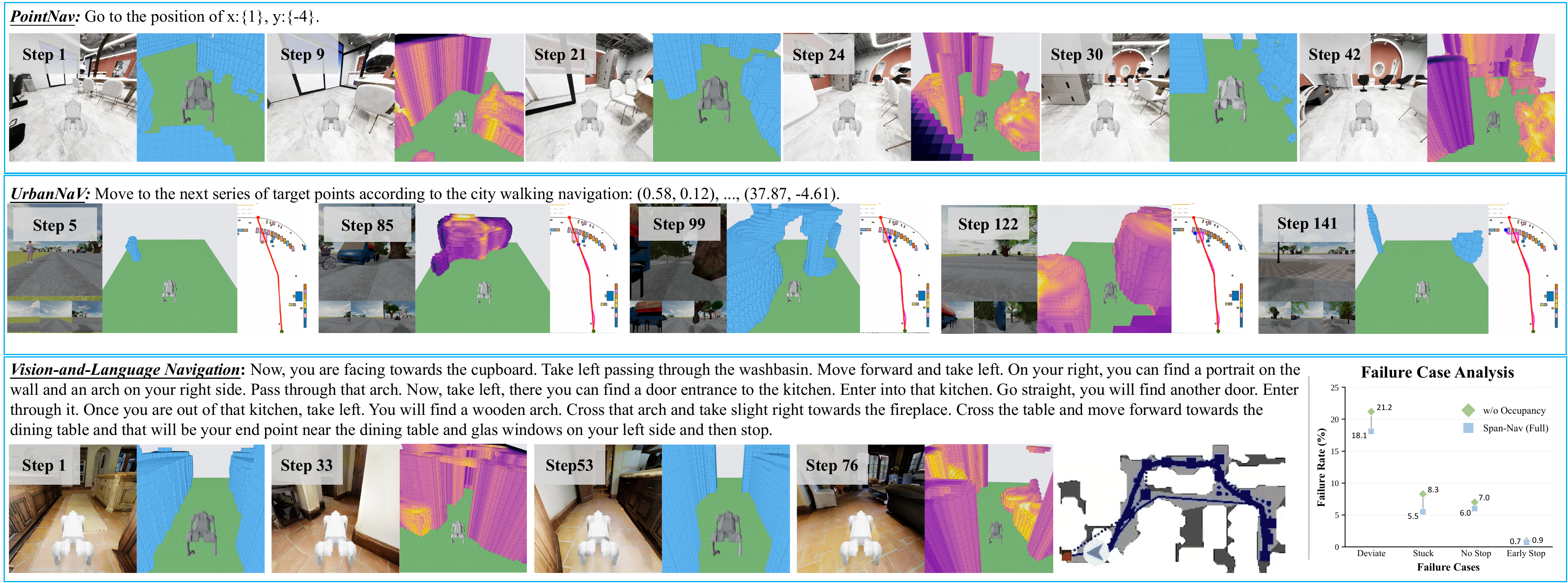}
  \vspace{-2mm}
  \caption{
    \textbf{Experimental visualization of \ours across diverse tasks in different simulators.}
    This includes PointNav, VLN, and UrbanNav, encompassing a variety of complex indoor and outdoor scenarios.
  }
  \label{fig:sim-vis}
  \vspace{-2mm}
\end{figure*}
\noindent\textbf{Navigation-Related Geometric Cues.}
The visualizations in \Cref{fig:sim-vis} further reveal how explicit geometric cues contribute to navigation. We visualize the predicted occupancy maps and the corresponding heat maps which reflect the importance of each occupancy voxel for action prediction across diverse tasks. The results indicate that the model consistently focuses more on geometry cues around the robot (highlight areas in \Cref{fig:sim-vis}), while the impact gradually decreases for regions farther away. This suggests that occupancy serves as the form of spatial awareness for action prediction, enabling the model to attend to navigation-related geometric cues such as nearby obstacles and traversable regions. The detailed heatmap computation method is provided in Appendix~\Cref{sec:app_sim}.
  

\begin{wrapfigure}[12]{r}{0.48\columnwidth}
\vspace{-1em}
  \centering
  \includegraphics[width=0.46\columnwidth]{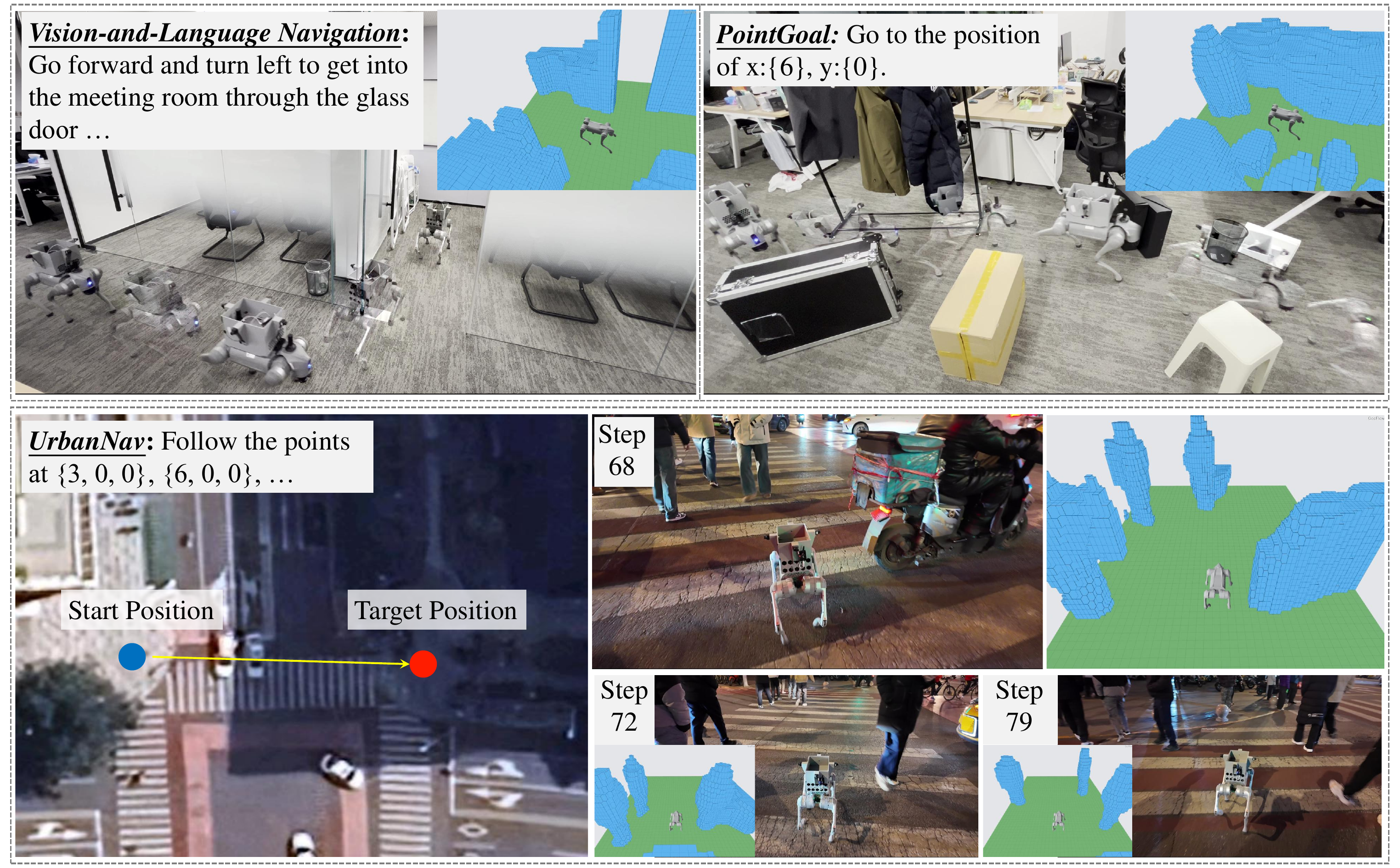}
  \caption{
    \textbf{Real-world experimental demonstration.} \ours demonstrates strong transferability.
  }
  \label{fig:real-vis}
\vspace{-1em}
\end{wrapfigure}
We further analyze the failure cases on RxR-CE in \Cref{fig:sim-vis}, categorizing them into four types: \textit{Deviate}, \textit{Stuck}, \textit{No Stop}, and \textit{Early Stop}. 
As shown in the results, \ours reduces the occurrence of both \textit{Stuck}, where the agent remains in place, and \textit{Deviate}, where the agent diverges from the ground-truth trajectory. 
These results indicate that \ours improves obstacle avoidance and enhances the agent's ability to recognize traversable space. 
More detailed failure-case analyses and additional visualizations are provided in the Appendix ~\Cref{sec:app_sim}.

\subsection{Real-World Results}
\begin{wraptable}{r}{0.58\linewidth}
    \centering
    \vspace{-1.5ex}
    \caption{\textbf{Deployment on one PointGoal task (10 trials).}}
    \label{tab:deployment}
    \vspace{0.3ex}

    \scriptsize
    \setlength{\tabcolsep}{3.5pt}
    \renewcommand{\arraystretch}{0.88}

    \resizebox{\linewidth}{!}{%
    \begin{tabular}{lcccc}
        \toprule
        Platform & Prec. & Infer. (ms) & E2E (ms) & Success \\
        \midrule
        Jetson AGX Orin   & FP16  & 860.3 & 957.6 & 2/10 \\
        Jetson AGX Orin   & INT8  & 794.4 & 889.1 & 1/10 \\
        RTX 4090 (remote) & FP16  & 190.1 & 349.9 & 7/10 \\
        Jetson AGX Thor   & NVFP4 & 147.8 & --    & -- \\
        \bottomrule
    \end{tabular}%
    }

    \vspace{-1.5ex}
\end{wraptable}
We further evaluate \ours in unseen real-world environments. As shown in \Cref{fig:real-vis}, \ours handles challenging cases such as transparent glass and fine-grained obstacle avoidance while producing meaningful occupancy predictions. These results demonstrate that occupancy-guided spatial awareness improves trajectory planning and supports more informed navigation decisions in complex physical environments. Our real-world evaluation covers 8 environments(6 indoor, 10 trials each; 2 outdoor, 3 trials each) using RGB only, without LiDAR. Originally, SPAN-Nav ran remotely on an RTX 4090, while the onboard Jetson AGX Orin handled sensing and control. Model inference takes 190.1 ms, with 159.8 ms system overhead (70.3 ms for capture/preprocessing and 89.5 ms for communication/control), yielding 349.9 ms E2E latency. ~\Cref{tab:deployment} further reports 8B onboard results: the high latency on Orin leads to more collision failures, whereas Thor with NVFP4 reduces inference latency to 147.8 ms. Although a Thor-equipped robot was unavailable, precluding E2E and success measurements, the result shows the potential of fully onboard inferenceSee more details for real world setting in Appendix \Cref{sec:app_real}.

\subsection{Ablation Study}
\begin{table*}[t]
\centering
\setlength{\tabcolsep}{0pt}

\begin{minipage}[t]{0.49\textwidth}
\vspace{0pt}
\centering
\caption{\textbf{Simulation results in InternVLA-N1 System-1 point-goal navigation benchmark.}}
\label{tab:sim_results}
\vspace{-2mm}
\scriptsize
\setlength{\tabcolsep}{3pt}
\renewcommand{\arraystretch}{1.08}
\resizebox{\linewidth}{!}{
\begin{tabular}{l|c|cc|cc}
\toprule
\multirow{2}{*}{Method} & \multirow{2}{*}{Obs.}
& \multicolumn{2}{c|}{Home}
& \multicolumn{2}{c}{Commercial} \\
& & SR$\uparrow$ & SPL$\uparrow$ & SR$\uparrow$ & SPL$\uparrow$ \\
\midrule
iPlanner~\citep{yang2023iplanner}$^\dagger$   & Depth & 43.0 & 40.6 & 54.6 & 52.8 \\
ViPlanner~\citep{roth2024viplanner}$^\dagger$ & RGB-D & 45.0 & 43.2 & 63.7 & 61.9 \\
LoGoPlanner~\citep{peng2025logoplanner}       & RGB-D & 57.3 & 52.4 & 67.1 & 63.9 \\
InternVLA-N1~\citep{wang2025internvla}(S1)    & RGB-D & 60.0 & 55.6 & 71.4 & 68.2 \\
MM-Nav(single)~\citep{xu2025mm}$^\dagger$         & RGB   & 79.9 & 77.6 & 86.7 & 84.7 \\
MM-Nav(multi)~\citep{xu2025mm}$^\dagger$         & RGB   & 86.3 & 81.1 & 89.9 & 85.5 \\
\rowcolor{rowcolor}
\textbf{\ours}$^\dagger$                      & RGB   & \textbf{90.9} & \textbf{85.7} & \textbf{91.0} & \textbf{88.1} \\
\bottomrule
\end{tabular}
}
\end{minipage}
\hfill
\begin{minipage}[t]{0.49\textwidth}
\vspace{0pt}
\centering
\caption{\textbf{Ablation study on architectural components and training strategies.}}
\label{tab:ablation_1}
\vspace{-2mm}
\scriptsize
\setlength{\tabcolsep}{3pt}
\renewcommand{\arraystretch}{1.08}
\resizebox{\linewidth}{!}{
\begin{tabular}{l|c|cc|cc}
\toprule
\multirow{2}{*}{Setting} & \multirow{2}{*}{IoU(\%)$\uparrow$}
& \multicolumn{2}{c|}{Home}
& \multicolumn{2}{c}{Commercial} \\
& & SR$\uparrow$ & SPL$\uparrow$ & SR$\uparrow$ & SPL$\uparrow$ \\
\midrule
discrete token          & 45.03          & 80.3 & 78.0 & 80.0 & 78.4 \\
w/o Occupancy           & -              & 70.5 & 67.9 & 75.1 & 73.1 \\
w/o Reasoning           & \textbf{60.71} & 81.3 & 77.2 & 85.2 & 82.3 \\
w/o Initialization      & 41.67          & 85.0 & 80.7 & 84.7 & 82.2 \\
w/o Training Stage I       & 46.84          & 85.7 & 78.5 & 88.5 & 83.9 \\
w/o Training Stage II      & 58.04          & 56.8 & 54.7 & 67.3 & 66.2 \\
\rowcolor{rowcolor} \textbf{\ours}          & 58.11          & \textbf{90.9} & \textbf{85.7} & \textbf{91.0} & \textbf{88.1} \\
\bottomrule
\end{tabular}
}
\end{minipage}

\vspace{1mm}
\begin{minipage}{0.98\textwidth}
\footnotesize
\textit{Note:} The symbol $^\dagger$ indicates zero-shot evaluation without additional fine-tuning on the Internscenes dataset.
\end{minipage}

\end{table*}

To systematically evaluate the contribution of each component, all ablations are conducted as single-factor removals, with identical training data and all other settings kept unchanged.

\noindent\textbf{Ablation analysis.}
We evaluate six architectural and training variants in \Cref{tab:ablation_1}. Discrete VQ-VAE tokens reduce occupancy IoU from 58.11 to 45.03, suggesting that quantization loses important spatial details in diverse navigation scenes. Removing occupancy supervision or spatial-prior-guided reasoning consistently lowers SR/SPL, confirming the importance of both spatial learning and its integration into action prediction. Removing encoder--decoder initialization drops IoU to 41.67 and degrades navigation performance, highlighting its role in stable spatial representation learning. 
The variant without Stage I supports the role of clean ground-truth priors
in stable spatial-to-action grounding. Removing Training Stage II causes the largest navigation decline despite a similar IoU (58.04 vs. 58.11), showing that Training Stage II mainly reduces train--test mismatch by aligning the policy with self-predicted spatial tokens. More detailed analysis is provided in Appendix \Cref{sec:app_sim}.

\noindent\textbf{Ablation on the number of spatial tokens.}

\begin{wrapfigure}{r}{0.48\columnwidth}
  \centering
  \vspace{-2.2em}
  \includegraphics[width=\linewidth]{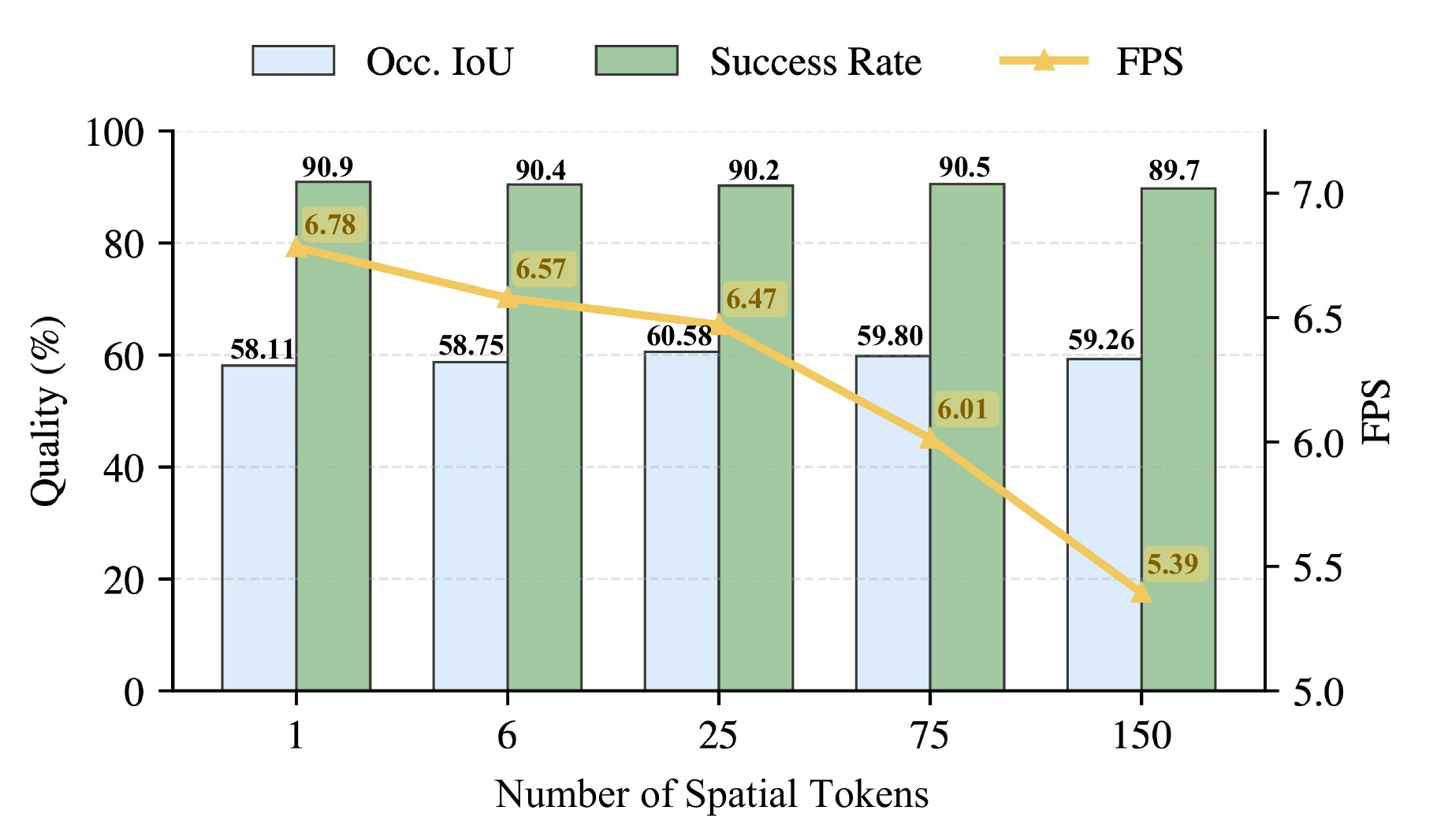}
  \caption{
    \textbf{Ablation study on the number of spatial tokens used for occupancy reconstruction.}
    We report occupancy reconstruction quality (IoU), success rate (SR) in point-goal navigation, and inference efficiency (FPS) under different spatial token budgets.
  }
  \label{fig:ablation_2}
  \vspace{-1em}
\end{wrapfigure}

We also vary the number of spatial tokens used for occupancy reconstruction and reasoning, as summarized in \Cref{fig:ablation_2}. When reducing the number of tokens from 150 to 1, occupancy IoU drops only marginally, while inference efficiency improves significantly, with a 26\% FPS increase using a single spatial token. Notably, the PointGoal success rate with one token is comparable to, or slightly higher than, larger token sets. This suggests that a compact single-token representation preserves navigation-critical spatial information while encouraging the model to distill task-relevant cues for downstream decision-making.


\section{Conclusion and Limitations} 
\label{sec:conclusion}

In this paper, we presented \ours, an end-to-end embodied navigation framework that learns generalized 3D spatial awareness through compact spatial token. By integrating spatial-prior-guided action reasoning, \ours effectively leverages 3D geometric cues for decision-making and achieves improved navigation performance and generalization across diverse tasks.
Despite these promising results, \ours has several limitations. First, it still relies on large-scale occupancy supervision, while high-quality real-world occupancy annotation requires costly and error-prone reconstruction pipelines. Second, the current evaluation uses limited camera and embodiment configurations, leaving cross-embodiment generalization across different views and robot sizes as future work. Third, in real-world deployment, network latency or unstable communication may reduce the policy execution frequency, leading to delayed actions and potential navigation failures.

\clearpage



\bibliography{references}  

\clearpage

\appendix

\begin{center}
    \LARGE\bfseries Appendix for \textcolor{spanGreen}{SP}\textcolor{spanTeal}{A}\textcolor{spanBlue}{N}-Nav: 
Generalized \\
\textcolor{spanGreen}{Sp}atial 
\textcolor{spanTeal}{A}ware\textcolor{spanBlue}{n}ess 
for Embodied Navigation
\end{center}

\section{Encoder and Decoder Initialization}
\label{sec:EDInit}
Following prior work on vector-quantized representation learning for occupancy reconstruction,
we employ a Vector-Quantized Variational Autoencoder (VQ-VAE) to pretrain the occupancy encoder
and decoder used in \ours.
The tokenizer has 51.33M parameters and is pretrained on 6.22M automatically collected occupancy-only samples. This is distinct from the 4.2M occupancy-annotated navigation samples within the 7.08M policy-training dataset. Each occupancy grid contains $120\times80\times20$ voxels of size $0.05\times0.05\times0.10$\,m, spanning 4\,m front, 2\,m rear, $\pm2$\,m laterally, and 2\,m vertically. The encoder produces $15\times10\times128$ features, with a 4,096-entry, 128-dimensional codebook.
Given a binary occupancy map $\mathbf{O}_t \in \{0,1\}^{H \times W \times D}$, we treat it as a 2D feature map
$\mathbf{x}_t \in \mathbb{R}^{D \times H \times W}$, where the height dimension is interpreted as the channel dimension.

A standard 2D ResNet--UNet style encoder extracts latent features $\mathbf{z}_t$, which are
vector-quantized into discrete code indices and quantized representations $\mathbf{z}_t^{\text{q}}$
via a learnable codebook.
A 2D decoder then reconstructs occupancy logits $\tilde{\mathbf{y}}_t \in \mathbb{R}^{D \times H \times W}$.

The VQ-VAE is trained with a multi-term objective:
\begin{align}
\mathcal{L}
= \lambda_{\text{c}}\mathcal{L}_{\text{c}}
+ \lambda_{\text{l}}\mathcal{L}_{\text{l}}
+ \lambda_{\text{vq}}\mathcal{L}_{\text{vq}},
\end{align}

where $\lambda_{\text{c}}, \lambda_{\text{l}}, \lambda_{\text{vq}}$ represent the hyperparameters for the occupancy reconstruction loss $\mathcal{L}_{\text{c}}$, Lovasz-hinge loss $\mathcal{L}_{\text{l}}$, and vector-quantization loss $\mathcal{L}_{\text{vq}}$, respectively, each of which is detailed in the subsequent discussion.

Let $o_i \in \mathbb{R}$ denote the predicted occupancy logit at voxel $i$ and
$y_i \in \{0,1\}$ is the corresponding ground-truth label.
The binary cross entropy loss for occupancy reconstruction is defined as
\begin{align}
\mathcal{L}_{c}
= - \frac{1}{N} \sum_{i=1}^{N}
\left[
y_i \log \sigma(o_i)
+ (1 - y_i)\log\left(1 - \sigma(o_i)\right)
\right],
\end{align}
where $\sigma(\cdot)$ denotes the sigmoid function and $N$ is the total number of voxels.

To better optimize the intersection-over-union (IoU) metric, we follow previous works and apply the Lovasz-hinge loss:
\begin{align}
\mathcal{L}_{l}
= \overline{\Delta}_{\mathrm{IoU}}
\bigl(
\mathrm{sort}(1 - o_i y_i)
\bigr),
\end{align}
where $\mathrm{sort}(\cdot)$ denotes sorting in descending order and
$\overline{\Delta}_{\mathrm{IoU}}(\cdot)$ is the Lovasz extension of the IoU loss.
The binary cross entropy loss and the Lovasz-hinge loss are also applied in the training of \ours.

To regularize the latent space, the vector-quantization module employs a standard codebook and a commitment loss $\mathcal{L}_{vq}$. We first optimize the VQ-VAE on a large-scale annotated occupancy dataset; the resulting encoder and decoder are subsequently leveraged to facilitate the end-to-end training of the full \ours framework.

\noindent\textbf{Optimization and downstream integration.}
VQ-VAE pretraining uses AdamW for one epoch, batch size 32, learning rate $5\times10^{-4}$, and weight decay 0.01, taking 20 hours on one RTX 4090. The encoder and decoder remain trainable in both navigation stages at learning rate $2\times10^{-5}$. The full model compresses the continuous $15\times10$ latent grid into one spatial token. The discrete-token ablation instead retains 150 VQ positions: code indices are predicted under cross-entropy supervision and their codebook embeddings are consumed by the action-reasoning pass. Occupancy predictions are decoded, thresholded at 0.5, and evaluated using mean occupied-voxel IoU.

\section{Trajectory Generation for Point Goal Navigation}
\label{sec:traj_for_point}
\subsection{\textbf{Overview}}
To generate collision-free navigation trajectories, we first discretize the $i$-th scene into a structured 3D representation. Specifically, we convert the scene into a set of $H$ height-sliced occupancy maps $\mathbf{O}_i=\{\mathbf{o}_{i,k}\}_{k=1}^{H}$, where each slice $\mathbf{o}_{i,k}$ encodes binary occupancy within the height interval $[k\Delta z,(k+1)\Delta z)$. We use a cell size of $0.05\,\mathrm{m}$ and a height interval $\Delta z=0.1\,\mathrm{m}$. To account for the robot’s physical dimensions, we fuse all slices below the robot height threshold ($z_{\text{limit}}=0.7\,\mathrm{m}$) into a 2D traversability map $\mathbf{O}_{\text{trav}}$, defined for each grid cell $(x,y)$ as
\begin{equation}
\mathbf{O}_{\text{trav}}(x,y)=\bigvee_{k=0}^{\lfloor z_{\text{limit}}/\Delta z \rfloor - 1}\mathbf{o}_{i,k}(x,y),
\end{equation}
where $\bigvee$ denotes an element-wise logical OR over height slices. $\mathbf{O}_{\text{trav}}(x,y)=1$ indicates the cell is occupied in at least one low-height slice and is thus non-traversable. Here $(x,y)$ denotes the planar coordinates of a grid cell in the world frame (with resolution $0.05\,\mathrm{m}$ per cell). Based on $\mathbf{O}_{\text{trav}}$, we derive two planning primitives: (i) a roadmap graph $\mathrm{G}=(\mathrm{V},\mathrm{E})$ constructed in free space and (ii) a distance-to-obstacle field $\mathrm{D}$ used as a clearance prior. In $\mathrm{G}$, $\mathrm{V}$ is the set of roadmap nodes (2D positions sampled in traversable free space) and $\mathrm{E}$ is the set of collision-free edges connecting nearby nodes.

On top of these primitives, we generate trajectories with an online \emph{A*+VO fusion} planner: A* provides global guidance on the roadmap, while a Velocity-Obstacles (VO) controller performs local collision avoidance and outputs executable motions. The resulting trajectory is a sequence of robot poses
\begin{equation}
\tau=\{(x_t,y_t,\mathrm{yaw}_t)\}_{t=0}^{T},
\end{equation}
where $(x_t,y_t)$ denotes the robot’s planar position in the world frame at time step $t$, and $\mathrm{yaw}_t$ is its heading angle (rotation about the vertical $z$-axis). In the following, we describe the design of the A* component and the VO component, and how they are fused into an efficient online planner.

\subsection{\textbf{Global Guidance via Roadmap A*}}
We run A* on a roadmap built in the known map constructed on the current robot's observation.
This produces a lightweight global guide that updates online as newly observed areas become available.


\textbf{Known Map Construction.}
We maintain a three-state known map $\mathbf{M}_{\text{known}}$ (\emph{free/occupied/unknown})
by accumulating the robot's observations over time (from [-2m,4m] in $x$-axis, and [-2m,2m] in $y$-axis).
During A* planning, unknown is treated as free to encourage optimistic exploration.
Every $K$ control steps, we filter the roadmap using the current $\mathbf{M}_{\text{known}}$ and re-run A* to obtain
an updated waypoint path $\tau^\star=\{q_m\}_{m=0}^{M}$, where each waypoint $q_m=(x_m,y_m)$ is a 2D position
in the world frame.

\textbf{A* Cost.}
To bias the global guide away from walls and narrow passages, we incorporate the distance-to-obstacle field $\mathrm{D}$ into the A* edge cost. 
For an edge $e=(i,j)\in \mathrm{E}$ connecting roadmap nodes $p_i$ and $p_j$, we define the cost
\begin{equation}
  w(e) = \|p_i-p_j\| + \lambda \max\!\left(0,\ d_{\text{pref}} - d_{\min}(e)\right),
\end{equation}
where $\|p_i-p_j\|$ is the Euclidean edge length, $d_{\min}(e)$ is the minimum value of $\mathrm{D}$ on the straight-line segment from $p_i$ to $p_j$,
and $d_{\text{pref}}$ is a user-defined preferred clearance threshold.
This cost favors short edges while adding extra penalty to edges whose clearance is below $d_{\text{pref}}$,
thereby encouraging A* to select routes that keep a larger safety margin from static obstacles.

\textbf{A*-guided Local Goal Selection.}
Let $\tau^\star=[q_0,\dots,q_M]$ denote the current A*-planned waypoint sequence,
where each waypoint $q_m=(x_m,y_m)$ is a 2D position on the roadmap.
Instead of tracking the next waypoint greedily, we select the \textbf{farthest} waypoint on $\tau^\star$ that is reachable by a straight line-of-sight segment as the VO local goal $g_{\text{local}}$. 
This choice reduces oscillations and provides a stable directional target for local control.

\subsection{\textbf{Local Avoidance via Truncated VO}}
 Given $g_{\text{local}}$ from A*, the VO controller chooses a short-horizon control that is
\emph{provably safe} (hard constraints) and \emph{preferentially high-clearance} (soft objective),
producing smooth motions while reacting to dynamic obstacles in real time.

\textbf{Robot modeling for VO.}
Since VO is formulated in velocity space, we approximate the robot footprint by an \emph{oriented rectangle}
of length $L$ and width $W$ (i.e., a rectangle whose orientation follows the robot heading $\mathrm{yaw}$).
Let $\{r^c\}_{c=1}^{4}$ be the four rectangle corners expressed in the robot frame.
For a candidate holonomic control $u=(v_x,v_y,\omega)$, where $(v_x,v_y)$ is the translational velocity in the robot frame
and $\omega$ is the yaw-rate command, the induced world-frame velocity of corner $c$ is approximated as
\begin{equation}
  v^c = R(\mathrm{yaw})\,[v_x, v_y]^\top + \omega \times r^c,
\end{equation}
where $R(\mathrm{yaw})$ rotates robot-frame velocities into the world frame and $\omega \times r^c$ is the tangential velocity
at the corner due to rotation about the vertical axis. A candidate control is accepted only if all corner velocities satisfy
the (truncated) VO constraints for dynamic obstacles, and the corresponding short-horizon rollout does not collide with the static occupancy map.

\textbf{Hard safety: truncated VO + lookahead checking.}
For each footprint corner $c$ and each dynamic obstacle, we construct a truncated velocity-obstacle (VO) constraint
with a finite horizon $T_{\mathrm{VO}}$. In this way we reject only candidate motions that may lead to collision within $T_{\mathrm{VO}}$
(instead of the infinite-horizon VO). Let $p^c$ be the corner position, and let the obstacle have center $p^o$ and velocity $v^o$.
We define the relative position and velocity as $p_{rel}^c=p^c-p^0$ and $v_{\mathrm{rel}}^c=v^c-v^o$.
A candidate is rejected if it is closing (with this definition, closing corresponds to $(p^c_{rel})^\top v_{\mathrm{rel}}^c<0$),
the relative-velocity direction lies inside the VO cone, and the relative speed is sufficient to reach the inflated obstacle within $T_{\mathrm{VO}}$.
In addition to the truncated VO , we perform a short-horizon forward rollout over $T_{\text{look}}$
and reject candidates whose predicted motion collides with the static occupancy map or with predicted dynamic obstacle positions.

\textbf{Soft preferences: tracking + smoothness + clearance.}
Among the safe candidates, we select the control $u=(v_x,v_y,\omega)$ minimizing the weighted objective:
\begin{equation}
\label{eq:vo_cost}
\begin{aligned}
J(u) &=
w_v\,\|v-v_{\mathrm{des}}\|
+ w_{curr}\,\|v-v_{\mathrm{curr}}\|
- w_{\mathrm{speed}}\,\|v\| \\
&\quad
+ w_\omega\,|\omega-\omega_{\mathrm{des}}|
+ J_{\mathrm{stat}}(d^{\mathrm{stat}}_{\min})
+ J_{\mathrm{dyn}}(d^{\mathrm{dyn}}_{\min}) \\
&\quad
+ w_a\,\|v\|(1-\hat v^{\top}\hat g)
,
\end{aligned}
\end{equation}
where all $w$'s are non-negative scalar weights
Here $v=[v_x,v_y]^\top$ is the translational command (robot frame), $v_{\mathrm{curr}}$ is the current translational velocity, and $(v_{\mathrm{des}},\omega_{\mathrm{des}})$ is the desired command derived from the local goal.
The first two terms promote tracking and smoothness, while $-w_{\mathrm{speed}}\|v\|$ mildly encourages forward progress.
The terms $w_\omega|\omega-\omega_{\mathrm{des}}|$ and $w_a\|v\|(1-\hat v^\top \hat g)$ favor smooth turning and goal-aligned motion, respectively.Here $\hat v$ is the normalized velocity direction, and $\hat g$ is the normalized direction from the robot to the local goal $g_{\text{local}}$ (both in the world frame).
We set $v_{\mathrm{des}}$ to be heading-aligned by default; if the heading error to $g_{\text{local}}$ is large, we set $v_{\mathrm{des}}=0$ to enable in-place turning and compute $\omega_{\mathrm{des}}$ from the heading error.

Let $d^{\mathrm{stat}}_{\min}$ be the minimum static clearance (queried from $\mathrm{D}$ along the rollout) after subtracting
a footprint margin $r_{\mathrm{eff}}$. We penalize low clearance only when it falls below a preferred threshold $d_{\mathrm{pref}}$:
\begin{equation}
J_{\mathrm{stat}} =
\begin{cases}
\displaystyle
\frac{w_c}{d^{\mathrm{stat}}_{\min}+\epsilon_c}, & d^{\mathrm{stat}}_{\min}<d_{\mathrm{pref}},\\[6pt]
0, & \text{otherwise},
\end{cases}
\end{equation}
where $w_c$ is the clearance-penalty weight and $\epsilon_c$ is a small constant.
Similarly, let $d^{\mathrm{dyn}}_{\min}$ be the minimum predicted clearance to dynamic obstacles over the lookahead horizon
(subtracting obstacle radius plus $r_{\mathrm{eff}}$). We penalize dynamic proximity more strongly:
\begin{equation}
J_{\mathrm{dyn}} =
\begin{cases}
\displaystyle
\frac{2w_c}{d^{\mathrm{dyn}}_{\min}+\epsilon_c}, & d^{\mathrm{dyn}}_{\min}<1.5\,d_{\mathrm{pref}},\\[6pt]
0, & \text{otherwise}.
\end{cases}
\end{equation}
Together, these terms make the controller naturally maintain a buffer from both static structures and moving agents, beyond merely avoiding collision.

\subsection{\textbf{Summary}}
Overall, the proposed A*+VO fusion planner enables efficient and reliable trajectory generation.
By combining A*-based global guidance with truncated-VO local avoidance and distance-map clearance priors,
it produces smooth, collision-free trajectories that naturally keep a safety buffer from obstacles.
Moreover, the trajectories are strongly aligned with the spatial reasoning process:
A* guidance is planned on a $\mathbf{M}_{\text{known}}$ that is incrementally built from the same receptive field used
to construct occupancy labels. 
Consequently, the explored/free/occupied structure that drives A* planning matches the occ labeling frame, yielding trajectories that are naturally consistent with our spatial prior guided action reasoning.
The A*+VO planner uses privileged simulator states during data collection, including information required for dynamic avoidance. This privilege is restricted to generating demonstrations. At inference, \ours receives RGB history and predicts current occupancy and trajectories in a closed loop. It does not explicitly track object identities or predict 4D scene flow; motion cues are implicit in the RGB history.

\section{Trajectory Generation for FrodoBots-2K Data}
\label{sec:traj_for_urban}
The FrodoBots-2K dataset was collected using low-cost onboard sensors, including wheel encoders and a magnetometer, resulting in limited observability and pronounced drift in the raw trajectories.
To obtain reliable action supervision, we perform 2D trajectory reconstruction using a dead-reckoning pipeline that exploits motion constraints and absolute heading observations.
Linear velocity is estimated from wheel encoder measurements, while zero-velocity updates (ZUPT) are applied during detected stationary intervals to enforce kinematic consistency and regularize motion estimation.
These zero-motion constraints improve the system's observability and are further used to suppress magnetometer-induced heading drift during low-dynamics periods.
Magnetometer measurements, after hard-iron calibration, provide absolute heading observations and are fused using a Kalman filter with a state vector $[yaw,\, yaw\_rate]$ under a constant angular velocity model.
To reduce accumulated drift and improve temporal consistency, Rauch--Tung--Striebel (RTS) smoothing is applied over the heading estimates.
Finally, the smoothed heading and regularized linear velocity are integrated to produce refined 2D trajectories, which serve as stable supervision for real-world navigation learning.

\section{Occupancy Annotation for Real World}
\label{sec:Occ_annotation}
We generate occupancy ground truth using a feed-forward depth reconstruction pipeline based on Depth Anything V3. Depth is jointly predicted from the current RGB frame and a temporally adjacent frame (1\,s later) to improve geometric stability under sparse viewpoints.
The predicted depth is converted into a 3D point cloud, from which ground and non-ground regions are separated via normal-based filtering and RANSAC plane fitting. Points whose surface normals are within $30^\circ$ of the vertical axis are treated as ground candidates.
The segmented point cloud is then voxelized using the same resolution, spatial extent, and coordinate conventions as the PointGoal Navigation benchmark. Occupancy labels are finally obtained through ray casting based on the camera pose and field of view.

\section{QA Loss and Two-Stage Training Strategy}
\label{sec:QA}
In the QA setting, the model leverages a vision-language architecture to predict answers derived from the provided image and query. We employ a standard cross-entropy loss to supervise the model, measuring the divergence between the predicted and actual labels as follows:

\[
\mathcal{L}_{\text{qa}} = - \sum_{i=1}^{N} \mathbf{y}_i \log(\text{P}(\hat{\mathbf{y}_i})),
\]
where $\mathbf{y}_i$ denotes the ground-truth answer in a one-hot encoded format, and $\mathbf{\hat{y}}_i$ represents the predicted probability distribution typically computed via a softmax activation function. Here, $\text{P}(\hat{y}_i)$ signifies the probability assigned by the model to each constituent token, with $N$ indicating the total number of words in the generated response.

In the two-stage training process, QA loss plays a critical role in improving the language understanding capacity of the model, as well as its ability to reason about the physical world.
In the first stage, we train the model using 2 million trajectories annotated with occupancy information and 0.5 million QA samples. In the second stage, we further train the model on the full set of trajectory and QA data. Through this two-stage training strategy, the model acquires generalized spatial perception capabilities, which subsequently enable accurate trajectory prediction.
Each stage runs for one epoch. We optimize with AdamW using global batch size 512, BF16 precision, and a cosine learning-rate schedule with 3\% warmup. The vision tower is frozen. The vision merger uses learning rate $2\times10^{-4}$; the VLM, action head, occupancy projectors, and VQ-VAE encoder/decoder use $2\times10^{-5}$. Full two-stage navigation training takes 12 hours on 32 H100 GPUs, in addition to the VQ-VAE pretraining cost above.

In particular, to evaluate the model’s ability to generalize across different agents, as discussed in \ours VLN-Occ-Free, we do not use occupancy annotations from the VLN data set at any stage of training.

\section{Real World Setting}
\label{sec:app_real}
\begin{figure}[t]
  \centering
  \includegraphics[width=0.5\textwidth]{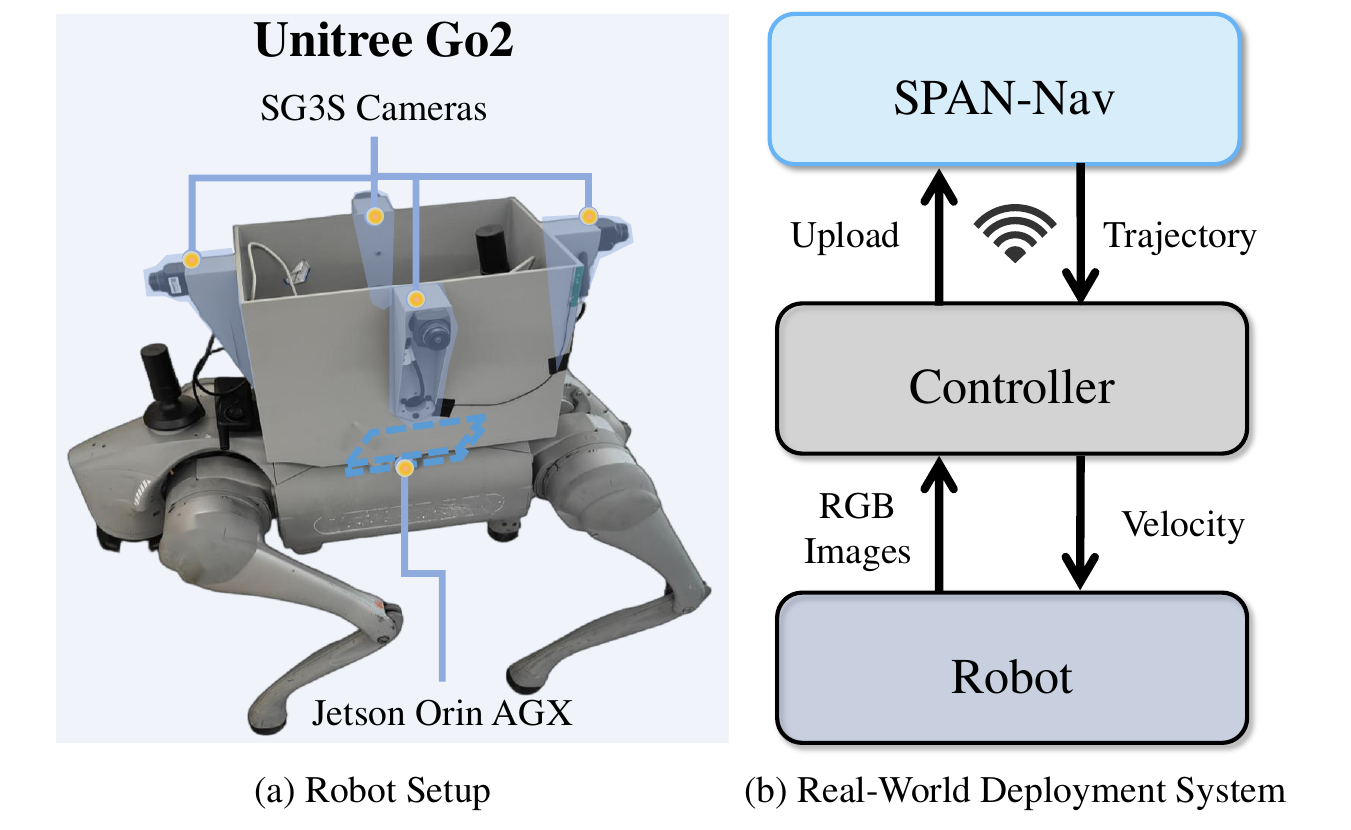}
  \caption{
    \textbf{Physical platform architecture.} (a) Computation unit (Jetson Orin AGX), and sensors. (b) Information flow and data pipeline during real-world deployment.
    }
  \label{fig:real-robot}
  
\end{figure}
Our real-world evaluation comprises eight environments: six indoor environments with 10 trials each and two outdoor environments with three trials each (66 trials in total). All reported navigation and obstacle-avoidance experiments use RGB only, without LiDAR. As illustrated in \Cref{fig:real-robot}, the model receives camera observations and language instructions and predicts local trajectories, which embodiment-specific controllers translate into velocity or joint commands.

\noindent\textbf{Remote and onboard inference.}
In the original setup, the model runs on a remote RTX 4090 and the onboard Jetson AGX Orin handles sensing and local control. Model inference takes 190.1\,ms; image capture/preprocessing adds 70.3\,ms and communication/control adds 89.5\,ms, for 159.8\,ms system overhead and 349.9\,ms end-to-end latency. We additionally compare the 8B model on one PointGoal task with 10 trials per robot setup. Orin FP16 and INT8 yield 2/10 and 1/10 successes, versus 7/10 for remote RTX 4090 FP16. Their higher end-to-end latency leads to more collision failures. Jetson AGX Thor achieves 147.8\,ms inference with NVFP4, but a Thor-equipped robot was unavailable; neither end-to-end latency nor navigation success was measured on Thor. These single-task counts are not aggregate success rates over the eight-environment evaluation.

\noindent\textbf{Embodiment transfer.}
The model predicts embodiment-agnostic trajectories executed by platform-specific controllers, and approximates robot footprint with an adjustable surround-view-camera bounding box. In principle, a larger platform such as Unitree B2 could use a camera rig whose bounding box encloses the body. This is a proposed deployment adaptation, not a demonstrated transfer result; differences in geometry, dynamics, and control remain to be evaluated.

\section{Simulation Results}
\label{sec:app_sim}

In this section, we provide additional details for simulation-based evaluation, including more qualitative visualizations, the heatmap visualization protocol used in \Cref{fig:sim-vis}, and a more detailed analysis of the ablation studies.

\subsection{Additional Simulation Visualizations}

\begin{figure*}[htpb]
  \centering
  \includegraphics[width=1.\textwidth]{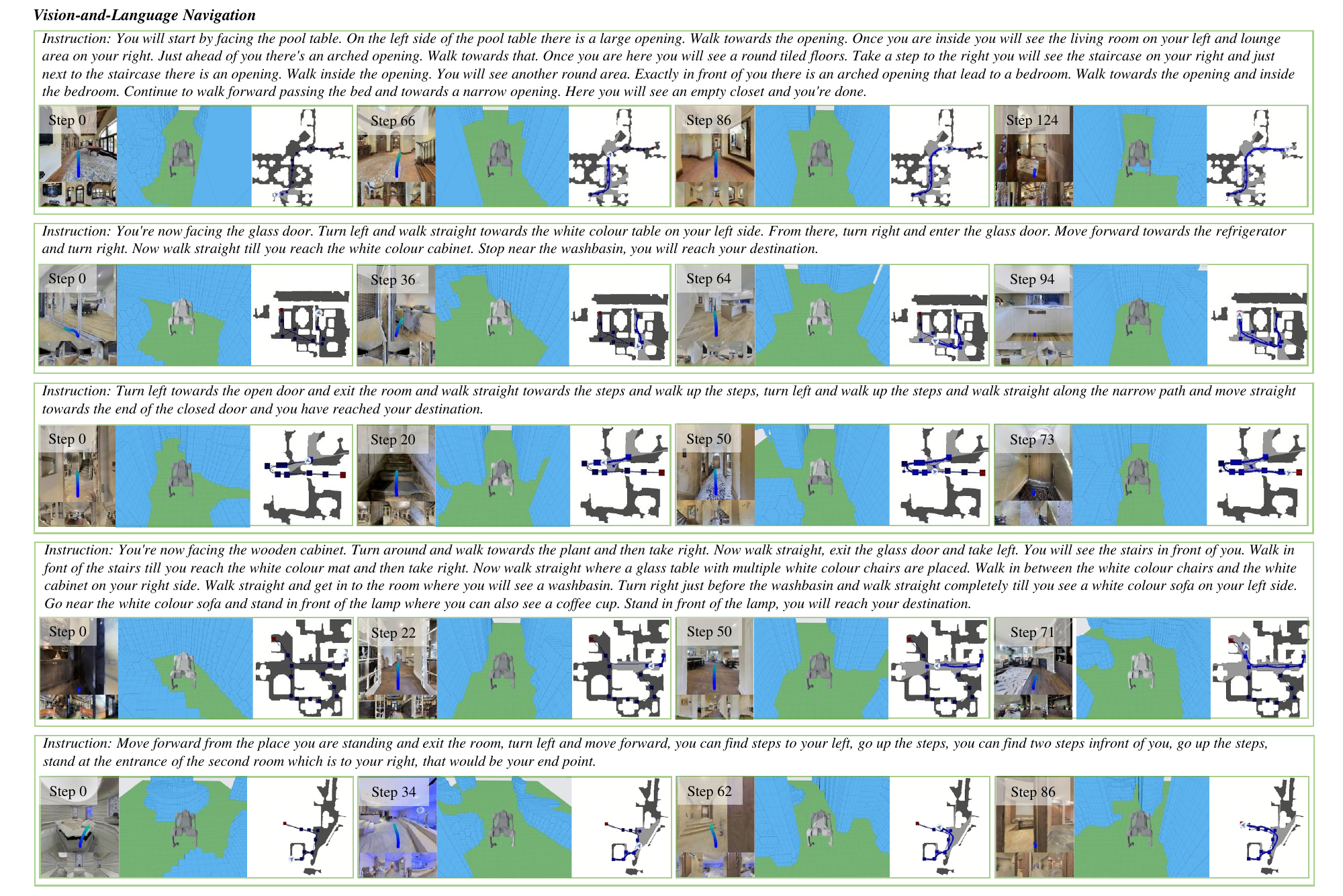}
  \caption{
    \textbf{Experimental Visualization of SPAN-Nav for VLN.}  
  }
  \label{fig:app-vln}
  
\end{figure*}

To further demonstrate the effectiveness of \ours, we provide additional experimental visualizations in unseen simulation environments. As shown in \Cref{fig:app-vln}, we present representative examples from Vision-and-Language Navigation (VLN) tasks. Under long-horizon instructions, the agent predicts plausible local occupancy maps that capture key traversable regions, such as doors, corridors, and intersections. By leveraging these explicit spatial cues, \ours performs more reliable spatial reasoning and executes navigation actions smoothly.

\begin{figure*}[htpb]
  \centering
  \includegraphics[width=1.\textwidth]{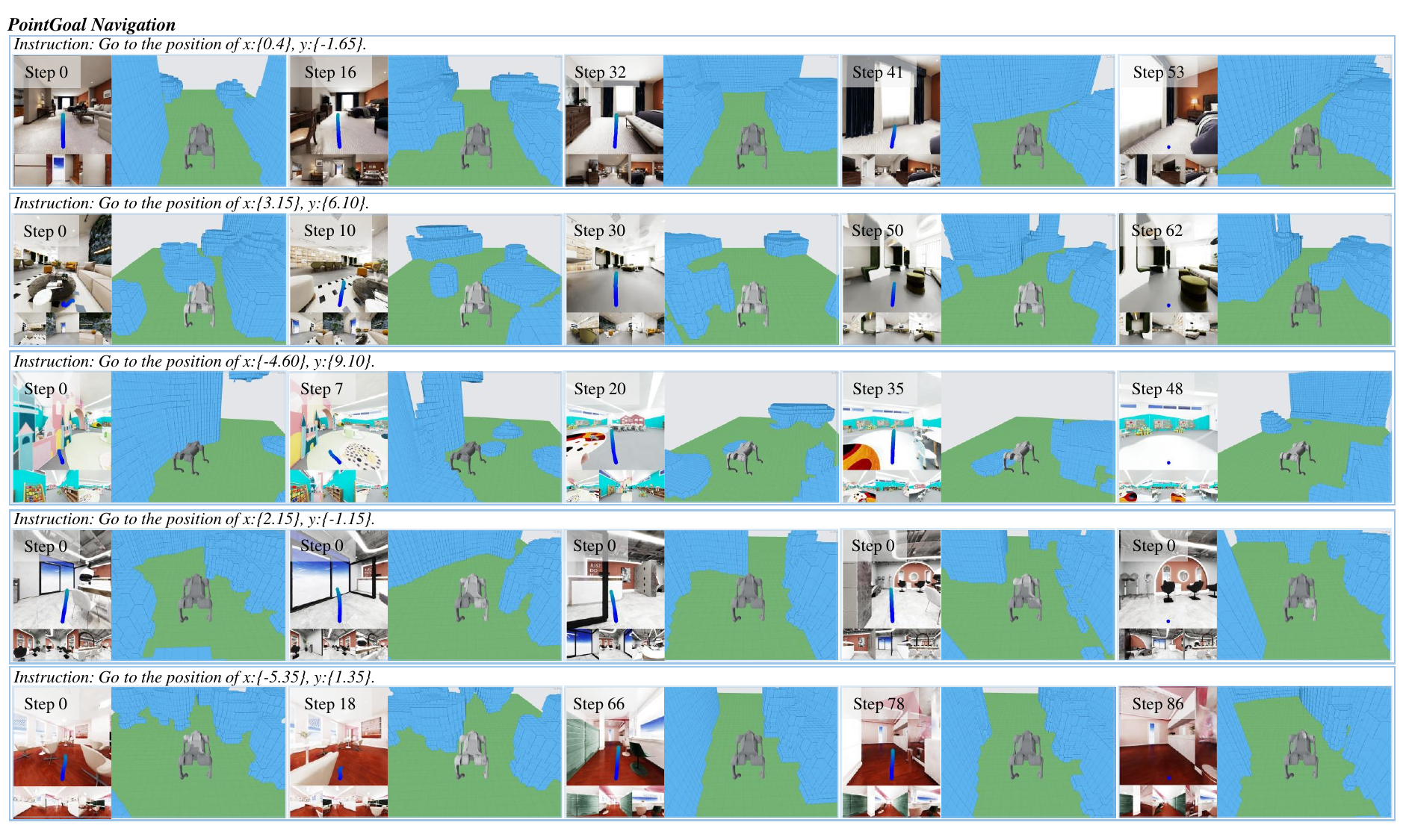}
  \caption{
    \textbf{Experimental Visualization of SPAN-Nav for PointGoal Navigation.}
  }
  \label{fig:app-pg}
  
\end{figure*}

\begin{figure*}[htpb]
  \centering
  \includegraphics[width=1.\textwidth]{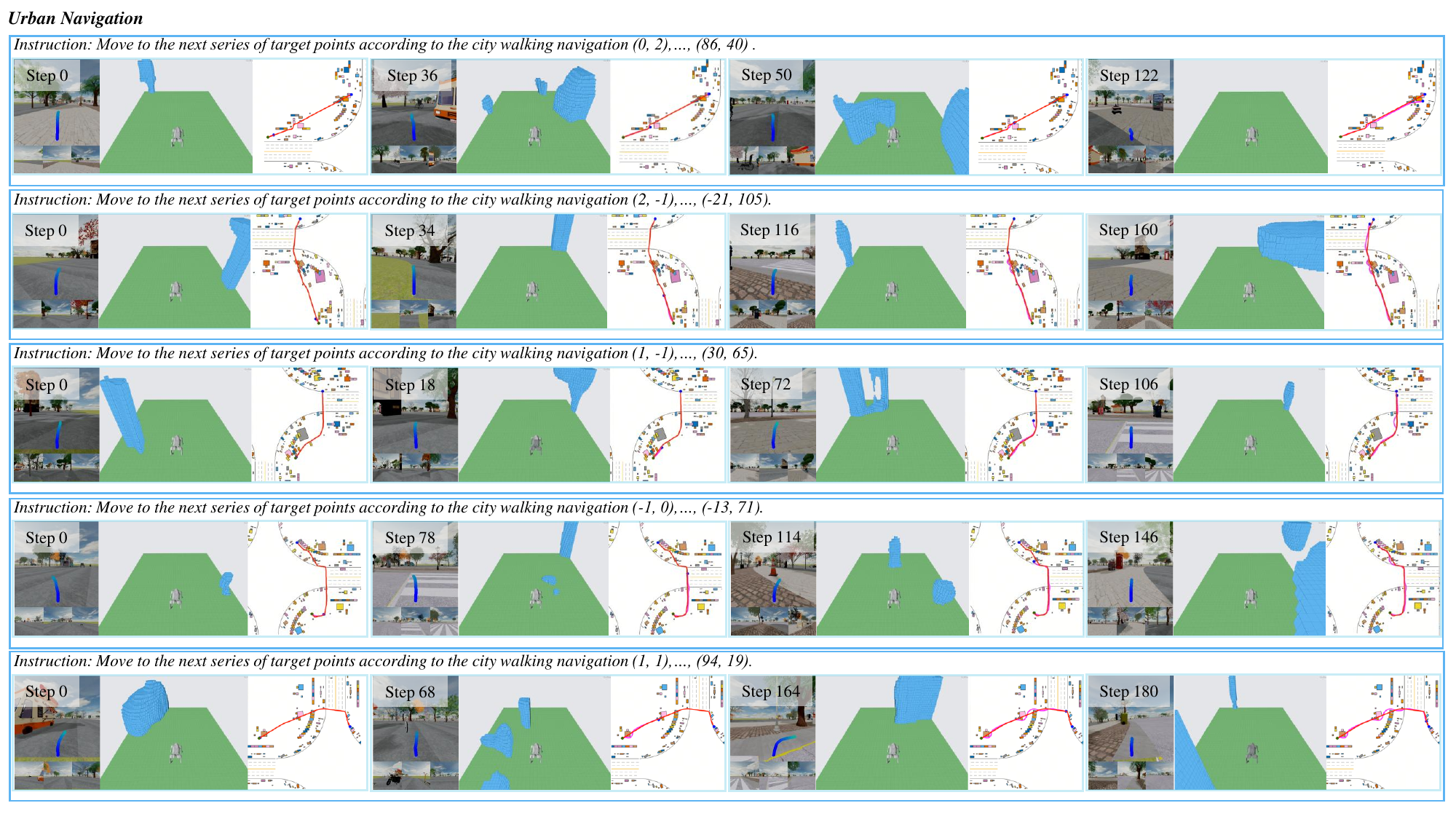}
  \caption{
    \textbf{Experimental Visualization of SPAN-Nav for Urban Navigation.}
  }
  \label{fig:app-ur}
  
\end{figure*}

Similarly, \Cref{fig:app-pg} shows the performance of \ours on PointGoal Navigation. Given a target point, the agent infers the 3D occupancy of surrounding obstacles and plans collision-free trajectories toward the goal. In addition, \Cref{fig:app-ur} presents qualitative results on Urban Navigation in outdoor simulation scenes. \ours effectively captures large-scale structural elements, including buildings, vegetation, and vehicles, which supports safer path planning and improves task completion in complex urban environments.

\subsection{Heatmap Visualization Protocol}

\noindent\textbf{Heatmap computation.}
We compute the heat maps in \Cref{fig:sim-vis} as a perturbation-based attribution over the predicted occupancy latent. Specifically, for each frame, we first run the standard \ours inference to obtain the predicted occupancy latent $z_t^{\text{out}}$, the decoded occupancy $\hat{O}_t=\mathrm{Dec}(z_t^{\text{out}})$, and the reference trajectory $\hat{T}_t=\{\hat{a}_m\}_{m=1}^{M}$. The occupancy latent $z_t^{\text{out}}$ can be viewed as a $\frac{H}{d}\times\frac{W}{d}$ grid of $C$-dimensional embeddings, where $d$ denotes the spatial downsampling factor of the occupancy encoder. We then slide a local $h\times w$ patch over this grid with a stride of $1$, and replace the latent vectors inside each patch with a noisy latent baseline. This perturbation measures how much the predicted action changes when a local spatial latent representation is corrupted.

For each perturbed patch $p$, we re-inject the modified occupancy latent $z_t^{p}$ into the action prediction branch and obtain a perturbed trajectory $\hat{T}^{p}_t=\{\hat{a}^{p}_m\}_{m=1}^{M}$. The patch-level importance score is computed by the normalized trajectory difference:
\[
s_p =
\left(
\sum_{m=1}^{M} w_m
\left\|
\frac{\hat{a}^{p}_m-\hat{a}_m}{\boldsymbol{\alpha}}
\right\|_2^2
\right)^{1/2},
\quad
w_m =
\frac{\exp(-(m-1)/\tau)}
{\sum_{n=1}^{M}\exp(-(n-1)/\tau)} ,
\]
where $\boldsymbol{\alpha}=(1.6,1.6,1.396)$ normalizes different action dimensions, and $\tau=2.0$ emphasizes near-term actions. A larger $s_p$ indicates that the corresponding occupancy latent patch $p$ has a stronger influence on the predicted trajectory.

Since latent patches are not one-to-one aligned with 3D voxels, we project this patch-level action influence back to the occupancy space using decoder responses. Specifically, we decode the perturbed latent to obtain $\hat{O}^{p}_t$, compute the voxel-wise change
\[
\Delta_p = |\hat{O}^{p}_t-\hat{O}_t|,
\]
and distribute the patch score according to the L1-normalized decoder difference:
\[
H_t(i,j,k)=
\sum_p
s_p
\frac{\Delta_p(i,j,k)}
{\sum_{i,j,k}\Delta_p(i,j,k)+\epsilon}.
\]
This normalization prevents patches with large diffuse decoder responses from dominating the heat map, while still assigning higher values to voxels that both respond to the latent perturbation and affect action prediction. The resulting heat values $H_t$ are rendered with $v_{\min}=0$ and $v_{\max}$ set to the 99th percentile of positive heat values. Therefore, brighter regions indicate occupancy voxels whose latent perturbation causes larger changes in the predicted trajectory, highlighting navigation-critical obstacles, traversable regions, and structural boundaries.

\subsection{Detailed Failure Cases Analysis}

We provide a more detailed analysis of the failure modes on the VLN-CE RxR val-unseen split to better understand how occupancy-induced spatial awareness affects navigation behavior. Following the analysis in the main paper, we categorize failed episodes into four types: \textit{Deviate}, \textit{Stuck}, \textit{No Stop}, and \textit{Early Stop}. \textit{Deviate} denotes cases where the agent moves away from the reference trajectory or enters an incorrect traversable region. \textit{Stuck} refers to cases where the agent fails to make effective progress, usually due to repeated collisions, oscillatory motions, or being trapped near obstacles. \textit{No Stop} indicates that the agent reaches the goal region but fails to issue the stop action, while \textit{Early Stop} corresponds to stopping before reaching the target.


\begin{wraptable}{r}{0.52\linewidth}
\vspace{-3mm}
\centering
\caption{\textbf{Effect of occupancy supervision on RxR Val-Unseen.}}
\label{tab:ablation_failure}
\scriptsize
\setlength{\tabcolsep}{2.4pt}
\renewcommand{\arraystretch}{0.92}
\begin{tabular}{l|c|c|c|c|c}
\toprule
    \textbf{Setting} & \textbf{SR$\uparrow$} & \textbf{Stuck$\downarrow$} & \textbf{Deviate$\downarrow$} & \textbf{No Stop$\downarrow$} & \textbf{Early Stop$\downarrow$} \\
\midrule
w/o Occupancy & 62.6 & 8.3 & 21.2 & 7.0 & 0.9 \\
\textbf{\ours~(Full)} & \textbf{69.7} & \textbf{5.5} & \textbf{18.1} & \textbf{6.0} & \textbf{0.7} \\
\bottomrule
\end{tabular}
\vspace{-4mm}
\end{wraptable}

As shown in the failure statistics \Cref{tab:ablation_failure}, removing occupancy supervision leads to more frequent geometry-related failures. Compared with the variant without occupancy prediction, the full \ours model improves SR from 62.6\% to 69.7\%. More importantly, the improvement mainly comes from reducing \textit{Stuck} and \textit{Deviate} failures. Specifically, the \textit{Stuck} rate decreases from 8.3\% to 5.5\%, and the \textit{Deviate} rate decreases from 21.2\% to 18.1\%. This suggests that explicit occupancy prediction helps the policy better understand nearby obstacles, free space, and feasible passages, rather than only serving as an auxiliary reconstruction objective.

The reduction in \textit{Stuck} failures indicates improved local obstacle awareness. 
Without occupancy supervision, the model mainly relies on RGB appearance and visual-language correlations, which can fail in cluttered regions, narrow passages, or door-like structures. 
\ours instead uses occupancy-supervised spatial tokens as an explicit spatial prior for trajectory generation, enabling more feasible waypoint prediction and reducing trapped behaviors.
\begin{figure*}[!t]
  \centering
  \includegraphics[width=1.\textwidth]{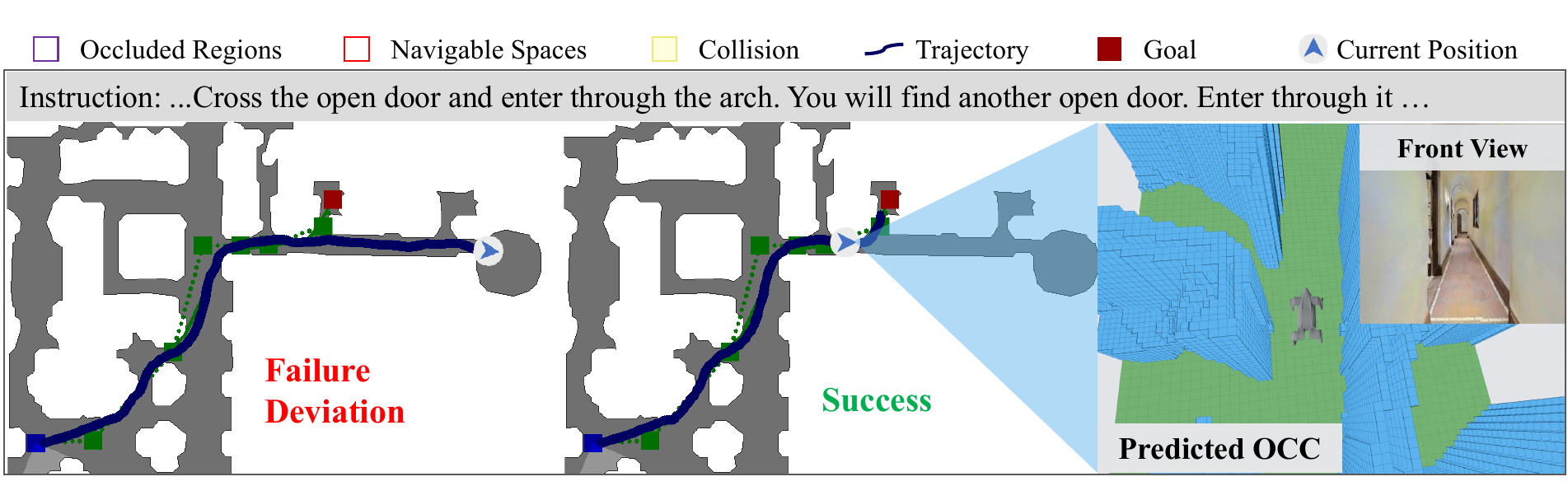}
  \caption{
  Visualization of a representative deviation failure case in VLN-CE. 
  }
  \label{fig:failure}
  
\end{figure*}

The decrease in \textit{Deviate} failures further demonstrates that occupancy prediction improves the recognition of traversable space. VLN instructions often require the agent to choose among multiple plausible paths, such as passing through an open doorway, following a corridor, or entering a specific room. Under partial observations, RGB-only policies may select a visually plausible but geometrically inconsistent route. As shown in ~\Cref{fig:failure}, by recovering local occupancy and plausible free space beyond partial observations, \ours can better distinguish accessible regions from blocked or irrelevant areas, thereby maintaining better consistency with the instructed route.

In contrast, the changes in \textit{No Stop} and \textit{Early Stop} are relatively smaller, decreasing from 7.0\% to 6.0\% and from 0.9\% to 0.7\%, respectively. This is expected because stopping behavior depends not only on geometric reasoning, but also on high-level instruction grounding, goal recognition, and progress estimation. 

Overall, the failure-case analysis shows that \ours mainly mitigates geometry-related navigation errors. 
By making local obstacles, traversable regions, and structural boundaries explicit for action prediction, occupancy supervision reduces \textit{Stuck} and \textit{Deviate} failures, supporting the role of occupancy-induced spatial awareness in robust embodied navigation.

\subsection{Detailed Ablation Analysis}

\noindent\textbf{Navigation-critical occupancy abstraction.}
By introducing spatial-prior-guided action reasoning, the compact spatial tokens learned by \ours focus more on navigation-critical occupancy cues rather than purely optimizing for pixel-wise reconstruction fidelity. For example, complex articulated structures such as swivel chairs and shelves are often represented as holistic non-traversable obstacles, instead of being reconstructed with fine-grained geometric details. This abstraction makes spatial reasoning and action planning easier, but may sacrifice some geometric fidelity, leading to a slight decrease in occupancy reconstruction IoU, as observed in the ablation study.

\noindent\textbf{Ablation on model architecture.}
To analyze the contribution of key architectural components, we compare \ours with three variants: (i) replacing continuous occupancy representations with discrete VQ-VAE tokens, denoted as \textit{Discrete Token}; (ii) removing occupancy supervision, denoted as \textit{w/o Occupancy}; and (iii) removing the spatial reasoning module, denoted as \textit{w/o Reasoning}. As shown in \Cref{tab:ablation_1}, using discrete VQ-VAE tokens significantly degrades occupancy reconstruction quality, with IoU dropping from 58.11 to 45.03. This indicates that quantized codebook representations struggle to capture diverse spatial layouts in embodied navigation, where spatial structures vary substantially and discretization may introduce irreversible information loss.

Removing occupancy supervision also causes a large performance drop, with SR / SPL decreasing to 70.5 / 67.9 on the Home split and 75.1 / 73.1 on the Commercial split. This highlights the importance of explicit occupancy supervision for learning spatial awareness and supporting downstream action planning. Removing the spatial CoT reasoning module leads to consistent declines in SR / SPL, dropping to 81.3 / 77.2 on the Home split and 85.2 / 82.3 on the Commercial split. These results suggest that explicit spatial reasoning helps the policy better utilize occupancy cues for action selection. Although \ours has a slightly lower IoU than the \textit{w/o CoT} variant, it achieves substantially better navigation performance, further indicating that spatial-prior-guided reasoning encourages the model to prioritize navigation-relevant occupancy cues over fine-grained geometric reconstruction.

\noindent\textbf{Ablation on training strategy.}
We investigate encoder--decoder initialization and both stages of S2A-Align. Without initialization, occupancy IoU drops to 41.67 and navigation degrades. The \textit{w/o Training Stage I} variant conditions action prediction on self-predicted spatial latents from the beginning of navigation training while preserving the pretrained tokenizer, training data, losses, and optimization steps. Its IoU is 46.84, with Home SR/SPL of 85.7/78.5 and Commercial SR/SPL of 88.5/83.9. Compared with the full model, this supports the role of Stage I in learning stable spatial-to-action grounding from clean ground-truth occupancy priors.

Removing Training Stage II causes a pronounced decline in navigation performance, with SR/SPL dropping to 56.8/54.7 on Home and 67.3/66.2 on Commercial, despite an occupancy IoU of 58.04, close to the full model's 58.11. Stage II therefore complements Stage I by adapting action reasoning to the self-predicted spatial priors encountered at inference and reducing the training--inference mismatch.


\end{document}